\documentclass{article}

\PassOptionsToPackage{numbers, compress}{natbib}

 \usepackage[main, final]{neurips_2026}

\usepackage[utf8]{inputenc} 
\usepackage[T1]{fontenc}    
\usepackage{hyperref}       
\usepackage{url}            
\usepackage{booktabs}       
\usepackage{amsfonts}       
\usepackage{nicefrac}       
\usepackage{microtype}      
\usepackage{xcolor}         
\usepackage{multirow}
\usepackage[table]{xcolor}
\usepackage{graphicx}
\usepackage{threeparttable}

\usepackage{hyperref}
\usepackage{url}

\usepackage{natbib}
\usepackage{graphicx}
\usepackage[utf8]{inputenc} 
\usepackage[T1]{fontenc}    
\usepackage{hyperref}       
\usepackage{url}            
\usepackage{booktabs}       
\usepackage{amsfonts}       
\usepackage{nicefrac}       
\usepackage{microtype}      
\usepackage{xcolor}         
\usepackage{amsmath}
\usepackage{enumitem}
\usepackage{subfig}
\usepackage{circledsteps}
\usepackage{amsthm}
\usepackage{wrapfig2}
\newtheorem{remark}{Remark}
\usepackage{hyperref}
\theoremstyle{plain}
\usepackage[textsize=tiny]{todonotes}
\usepackage{amsmath,amssymb}

\newtheorem{theorem}{Theorem}
\newtheorem{proposition}{Proposition}
\newtheorem{corollary}{Corollary}

\title{ReCoG: Reciprocal Co-Evolution for \\ Multimodal Graph Learning}

\author{%
  Rui~Xue \& Tianfu~Wu \\
Department of Electrical and Computer Engineering (ECE)\\
North Carolina State University, Raleigh, NC 27695, USA \\
\texttt{\{rxue,twu19\}@ncsu.edu} 
}

\begin{document}

\maketitle

\begin{abstract}

Multimodal graph learning requires jointly training over graph structure and heterogeneous node attributes, yet existing methods largely decouple these processes: prior multimodal graph neural networks (GNNs) focus on aligning modalities in a shared embedding space while operating on fixed or weakly adapted graph structures, and graph structure learning approaches infer topology from unimodal node representations without accounting for multimodal interactions. This separation fundamentally limits the ability of GNNs to capture semantically meaningful relationships in multimodal settings, where observed edges are often noisy, incomplete, or misaligned with underlying semantics.
We propose \textbf{ReCoG} (Reciprocal Co-Evolution for Multimodal Graph Learning), a new learning paradigm that tightly couples graph structure learning and multimodal representation learning through \textbf{end-to-end reciprocal interaction}.
Concretely, ReCoG integrates (i) \textbf{a multimodal graph refiner} that infers and corrects edges using cross-modal semantic evidence, and (ii) \textbf{a coupled cross-modal message passing mechanism} that performs joint intra- and inter-modality propagation over the refined graph. This unified design yields 
greater expressiveness than decoupled or two-stage formulations and allows dynamic interaction between topology and representation learning.
Across diverse benchmarks for node classification and link prediction, ReCoG consistently outperforms strong multimodal graph structure learning baselines, including graph foundation models. Our results demonstrate that \textbf{reciprocal co-evolution of structure and semantics is important for effective multimodal graph learning}, challenging the prevailing separation between topology and representation learning.
\end{abstract}

\setlength{\abovecaptionskip}{1pt}
\setlength{\belowcaptionskip}{-6pt}
\setlength{\textfloatsep}{10pt}   
\setlength{\intextsep}{10pt}      
\setlength{\belowdisplayskip}{1pt} \setlength{\belowdisplayshortskip}{1pt}
\setlength{\abovedisplayskip}{1pt} \setlength{\abovedisplayshortskip}{1pt} 

\section{Introduction}
\label{sec:intro}
\vspace{-2mm}
 
Graph neural networks (GNNs) have become a dominant paradigm for learning over relational data, achieving strong performance across applications such as recommendation systems, social networks, and scientific modeling~\citep{tang2020knowing,sankar2021graph,fout2017protein,wu2022graph,zhang2024linear,kipf2016semi,gasteiger2018combining,velivckovic2017graph,wu2019simplifying,xue2023lazygnn,10.1145/3580305.3599565}. While most existing GNNs assume \textit{unimodal} node attributes, real-world graphs are inherently \textbf{multimodal}, where nodes are described by heterogeneous signals such as text, images, and metadata. These modalities provide complementary semantic views, making multimodal graph learning an important step toward more realistic and expressive representation learning.
 
Despite recent progress, existing approaches to multimodal graph learning remain fundamentally limited by a \textbf{decoupled design}. On one hand, multimodal GNNs typically focus on aligning representations across modalities—often through separate encoders or late fusion—while operating on \textbf{fixed or weakly adapted graph structures}. 
On the other hand, 
prior work on graph structure learning (e.g., IDGL~\cite{chen2020idgl}, SLAPS~\cite{fatemi2021slaps}, Pro-GNN~\cite{jin2020prognn}, NodeFormer~\cite{wu2022nodeformer}) iteratively refines topology jointly with representations, but operates in the unimodal setting and does not model cross-modal evidence in either the refinement signal or the propagation rule.
As a result, these methods fail to capture the tight interplay between structure and semantics that is essential in multimodal settings, where observed edges are often noisy, incomplete, or misaligned with multimodal semantic relationships.

\begin{figure*}[t]
\centering
\includegraphics[width=0.97\textwidth]{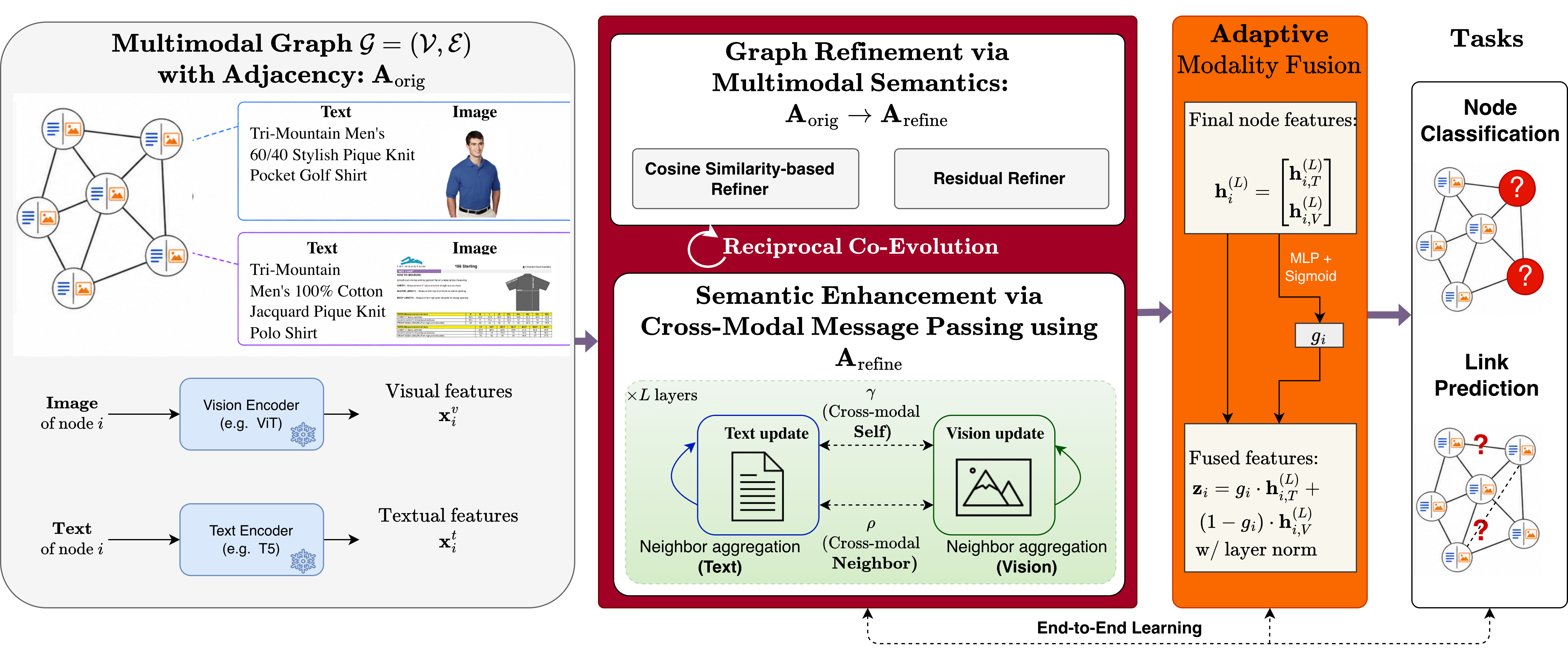}
\caption{Overview of the proposed ReCoG framework. 
\textbf{Left:} The input multimodal graph with text and vision features. 
\textbf{Center-top:} The Structure Refiner module with two variants: 
a similarity-based version, and 
a residual MLP-based version. 
\textbf{Center-bottom:} Cross-modal coupled message passing over $L$ layers, where each layer performs intra-modal neighbor aggregation and cross-modal injection.
\textbf{Right:} Modality fusion followed by a task-specific head. 
The entire pipeline is trained end-to-end.
\textit{Best viewed in magnification.} See text for detail.
}
\label{fig:framework}
\end{figure*}

We argue that \textbf{graph structure and multimodal representations} should not be treated as independent components, but rather as \textbf{mutually dependent variables that must be learned jointly}. 
In multimodal graphs, better representations provide stronger semantic signals for identifying meaningful connections, while improved graph structure enables more effective cross-modal information propagation. 

To realize this idea, we propose \textbf{ReCoG} (Reciprocal Co-Evolution for Multimodal Graph Learning, Fig.~\ref{fig:framework}), a unified framework that tightly couples graph structure learning and multimodal representation learning through \textbf{end-to-end reciprocal interaction}. ReCoG consists of two interdependent components:
\begin{itemize}[leftmargin=*,noitemsep,topsep=0pt]
    \item \textbf{A multimodal graph refiner} leverages cross-modal semantic signals to denoise and augment the observed topology, correcting spurious edges and recovering missing but semantically meaningful connections (Fig.~\ref{fig:refinement} and Appendix~\ref{app:misalign}). 
    \item \textbf{A cross-modal message passing mechanism} performs joint intra- and inter-modality propagation over the evolving graph, enabling fine-grained interaction between modalities at every layer. 
\end{itemize}
Importantly, these components are optimized jointly: updated representations iteratively guide topology refinement, while the refined graph directly shapes subsequent cross-modal propagation.

We evaluate ReCoG on nine multimodal graph benchmarks spanning node classification and link prediction tasks. Across diverse encoder configurations, ReCoG consistently outperforms strong multimodal, and graph foundation model baselines, often by significant margins. 
These results demonstrate that \textbf{jointly evolving graph structure and multimodal semantics is important for effective multimodal graph learning.}
 
\textbf{Our contributions.} We provide details of related work positioning our proposed ReCoG in Appendix~\ref{app:related}. It makes three main contributions as follows:
\begin{itemize}[leftmargin=*,noitemsep,topsep=0pt]
    \item  \textbf{A unified multimodal graph learning framework.} We propose ReCoG, which integrates a multimodal graph refiner with a coupled cross-modal message passing mechanism, enabling iterative interaction between structure and semantics throughout training, and challenging the conventional separation between topology and feature learning.
    \item \textbf{Spectral and contraction analysis of coupled propagation.} We give a supergraph reformulation of the coupled update that admits a closed-form spectral characterization of the modality coupling coefficients, and prove a depth-wise contraction bound that motivates our shared-weight design.
    \item  \textbf{Strong empirical performance.} We demonstrate consistent improvements over state-of-the-art multimodal GNNs and graph structure learning methods across nine benchmarks for node classification and link prediction.
\end{itemize}

\section{Methodology}
\label{sec:method}
 \vspace{-2mm}
We now describe the three components of ReCoG (Fig.~\ref{fig:framework}): (1) a multimodal graph refiner that refines the observed topology (Sec.~\ref{sec:refiner}), (2) a cross-modal message passing module that propagates information over the refined graph (Sec.~\ref{sec:gnn}), and (3) an adaptive modality fusion layer that integrates modality-specific representations (Sec.~\ref{sec:fusion}). These components are optimized end-to-end, forming a closed loop between topology refinement and representation learning.

\subsection{Graph Refinement via Multimodal Semantics}
\label{sec:refiner}
\vspace{-2mm}

We refine the observed topology using multimodal semantic evidence that adapts dynamically as representations evolve. We first state the refinement problem in general, then present two complementary instantiations: a \emph{cosine similarity refiner} that is efficient and interpretable, and a \emph{residual refiner} that provides greater expressive power.

\textbf{Problem formulation.}
Given the multimodal graph $\mathcal{G} = (\mathcal{V}, \mathcal{E})$ with adjacency $\mathbf{A}_{\text{orig}}$ and multimodal node features $\{\mathbf{x}_i^t, \mathbf{x}_i^v\}_{i \in \mathcal{V}}$, the refiner produces a refined adjacency $\mathbf{A}_{\text{refine}} \in [0,1]^{|\mathcal{V}| \times |\mathcal{V}|}$ by combining the original topology with a learned multimodal edge score.

\textbf{Mini-batch target-centric candidates.}
Large-scale GNN training typically uses neighbor sampling. Let $\mathcal{V}_{\mathcal{B}}$ denote the sampled subgraph and $\mathcal{B} \subseteq \mathcal{V}_{\mathcal{B}}$ the set of target nodes, with $|\mathcal{B}|=B$ and $|\mathcal{V}_{\mathcal{B}}|=N_{\mathcal{B}}$. Since only target nodes receive direct supervision~\citep{xue2023efficient}, we restrict refinement to target-centric pairs:
\begin{equation}
    \mathcal{P} = \big\{(i,j) : i \in \mathcal{B},\; j \in \mathcal{V}_{\mathcal{B}}\setminus\{i\}\big\}
    = \mathcal{S}_{\mathcal{E}} \cup \mathcal{S}_{\text{cand}},
\end{equation}
where $\mathcal{S}_{\mathcal{E}} \subset \mathcal{E}$ contains original edges to be re-scored (\emph{denoising}) and $\mathcal{S}_{\text{cand}}$ ($\mathcal{S}_{\text{cand}} \cap \mathcal{E} = \emptyset$) contains candidate non-edges to be discovered (\emph{recovery}). This reduces candidate complexity from $O(N_{\mathcal{B}}^2)$ to $O(B N_{\mathcal{B}})$ We provide a full analysis in Appendix~\ref{app:complexity}.

\subsubsection{Cosine Similarity Refiner}
\vspace{-2mm}
This variant provides a simple and interpretable mechanism based on pretrained feature geometry.

\textbf{Multimodal similarity scoring.}
We compute modality-specific cosine similarities and combine them via a learnable convex combination to obtain $s(i,j)$.
\begin{equation}    
    s(i,j) = \lambda\, S_V(i,j) + (1-\lambda)\, S_T(i,j),
     \text{ with } S_T(i,j) = \cos(\mathbf{x}_i^t, \mathbf{x}_j^t),
    \,
    S_V(i,j) = \cos(\mathbf{x}_i^v, \mathbf{x}_j^v),
    \label{eq:sim}
\end{equation}
where $\lambda = \sigma(\lambda_{\text{logit}}) \in (0,1)$ is learned. Textual similarity captures contextual alignment, visual similarity captures perceptual resemblance, and $\lambda$ lets the model balance the two on a per-task basis. We mask self-loops and clamp negative similarities to zero
to prevent anti-semantic edges.

\textbf{Semantic neighbor selection.}
For each target node $i\in \mathcal{B}$, we select the top-$k$ neighbors by similarity:
\begin{equation}
    \mathcal{N}_{\text{sem}}(i) = \text{TopK}_{j\in \mathcal{V}_{\mathcal{B}}} \{ s(i,j) \}, \qquad |\mathcal{N}_{\text{sem}}(i)| = k.
    \label{eq:sim_topk}
\end{equation}
where $k$ is a hyperparameter in training. This produces a set of candidate semantic edges, each associated with a similarity score. And we set $s(i,j)=0, j\notin \mathcal{N}_{\text{sem}}(i)$.

\textbf{Topology integration.}
We combine original and semantic edges via a convex combination:
\begin{equation}
\mathbf{A}_{\text{refine}}(i,j) = \text{Clamp}\Big(\alpha\cdot \mathbf{A}_{\text{orig}}(i,j) + (1-\alpha)\cdot s(i,j), 0, 1\Big), \forall (i,j)\in \mathcal{P},
\label{eq:adj_sim}
\end{equation}
where $\alpha = \sigma(\alpha_{\text{logit}}) \in (0,1)$ is a learnable coefficient controlling the trade-off between observed topology and semantic edges. This design preserves the original graph as a structural prior while introducing new edges only when supported by strong multimodal similarity. However, because $s(i,j) \geq 0$, this rule can only \emph{add or strengthen} edges relative to $\mathbf{A}_{\text{orig}}$.

\subsubsection{Residual Refiner}
\vspace{-2mm}
The cosine similarity refiner relies on a fixed functional form and collapses feature interactions into a scalar. To capture richer, task-dependent relationships, we introduce a more expressive \emph{residual refinement} mechanism.

\textbf{Pairwise feature and residual scoring.}
For each pair $(i,j)$, we construct a pairwise feature that preserves dimension-wise interaction patterns:
\begin{equation}
\varphi(i,j) =
[\mathbf{x}_i^t,\mathbf{x}_j^t,
\mathbf{x}_i^v,\mathbf{x}_j^v,
\mathbf{x}_i^t\odot\mathbf{x}_j^t,
\mathbf{x}_i^v\odot\mathbf{x}_j^v,
|\mathbf{x}_i^t-\mathbf{x}_j^t|,
|\mathbf{x}_i^v-\mathbf{x}_j^v|],
\label{eq:phi}
\end{equation}
combining individual node features, multiplicative interactions, and modality-specific differences rather than collapsing them into a scalar. We pass the pairwise feature through a learned MLP scorer $f_\theta$ to produce a \emph{residual edge score} $r(i,j)\in(-1,1)$, which indicates the strength and direction of the predicted topological correction:
\begin{equation}
r(i,j) = \tanh\big(f_\theta(\varphi(i,j))\big). 
\label{eq:residual-rewrite}
\end{equation}
A positive $r(i,j)$ encourages edge addition or strengthening, while a negative value signals edge removal or weakening.

\textbf{Coarse-to-fine candidate selection.}
The complexity $O(BN_{\mathcal{B}})$ can be still expensive when $N_{\mathcal{B}}$ is large in conjuction with the high-dim of $\varphi(i,j)$. We therefore adopt a coarse-to-fine strategy:
\textbf{(i) Coarse filtering.}
We project each node into a low-dimensional embedding:
$
\mathbf{h}_i = \mathbf{W}_c [\mathbf{x}_i^t ; \mathbf{x}_i^v],
$
where $\mathbf{W}_c \in \mathbb{R}^{d_h \times (d_t + d_v)}$ and $d_h \ll d_t + d_v$.
For each target node $i \in \mathcal{B}$, we compute the cosine similarity $\tilde{S}(i,j) = \cos(\mathbf{h}_i, \mathbf{h}_j)$ over all $j \in \mathcal{V}_{\mathcal{B}}$ and retain the top-$c$ candidates, where $c = \eta k$ with overselection ratio $\eta > 1$.

\textbf{(ii) Fine scoring.}
We apply the MLP scorer to candidate pairs $(i,j)$ with $j \in \mathcal{C}(i)$ and select the final top-$k$ neighbors. Formally:
\begin{equation}
\textbf{Coarse:} \quad\mathcal{C}(i) = \text{TopK}_{j\in \mathcal{V}_{\mathcal{B}}} \, \tilde{S}(i,j), \quad
\textbf{Fine:}\quad\mathcal{N}_{\text{sem}}(i) = \text{TopK}_{j \in \mathcal{C}(i)} r(i,j). \label{eqn:res_topk}
\end{equation}

\textbf{Topology integration.}
We then integrate the MLP-based scores into the original topology as a residual signal:
\begin{equation}
\mathbf{A}_{\text{refine}}(i,j) =
\text{Clamp}\big(
\mathbf{A}_{\text{orig}}(i,j) + \delta \cdot r(i,j),\; 0,\;1
\big), \label{eq:adj_res}
\end{equation}
where $\delta = \sigma(\delta_{\text{logit}}) \in (0,1)$ is a learnable scalar that controls the refinement magnitude. Because $r(i,j)$ is signed, this rule supports both \emph{edge denoising} and \emph{semantic edge recovery} — a strict capability gain over Eq.~\eqref{eq:adj_sim}.
Proposition~\ref{prop:refiner} shows that the family of $\mathbf{A}_{\text{refine}}$ functions induced by Eq.~\eqref{eq:adj_res} strictly contains those induced by Eq.~\eqref{eq:adj_sim}.

\begin{proposition}[\textbf{Strict Generalization of Residual Refinement}]
\label{prop:refiner}
On a fixed candidate set $\mathcal{C} \subseteq V \times V$, let $\mathcal{F}_{\mathrm{sim}}$ denote the family of edge-scoring functions induced by the similarity refiner (Eqn.~\eqref{eq:sim}), and let $\mathcal{F}_{\mathrm{res}}$ denote the family induced by the residual refiner (Eq.~\eqref{eq:residual-rewrite}). Assume the similarity refiner's correction magnitude is bounded by $\delta$ on $\mathcal{C}$. Then $\mathcal{F}_{\mathrm{sim}} \subsetneq \mathcal{F}_{\mathrm{res}}$. 
\end{proposition}

\begin{remark}
\label{app:remark1}
Proposition~\ref{prop:refiner} characterizes \emph{expressiveness}, 
not generalization. The inclusion follows from a universal-approximation 
argument on the residual MLP, and strictness from dimension-wise 
interactions (e.g., $\mathbf{x}_i^t \odot \mathbf{x}_j^t$) that 
cosine similarity collapses to a scalar. Whether this additional capacity 
translates to better task performance depends on the learning context: 
Sec.~\ref{sec:ablation} show that the residual 
refiner's advantage emerges only when coupled with cross-modal propagation.
\end{remark}

\begin{wrapfigure}{r}{0.55\textwidth}\vspace{-12mm}
    \centering
    \includegraphics[width=\linewidth]{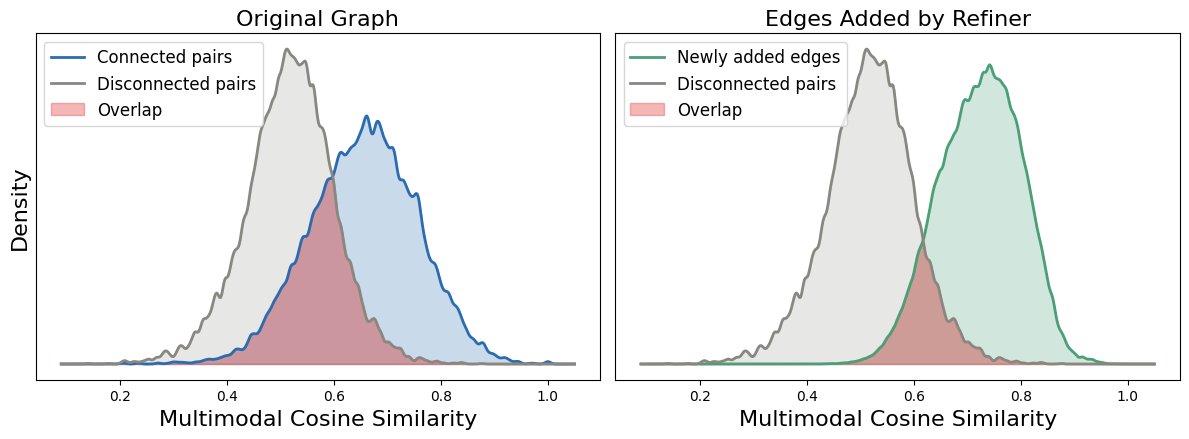}
    \caption{
    \textbf{Topology–semantics alignment before and after refinement.}
    \textbf{Left}: distribution of multimodal cosine similarity for connected and disconnected node pairs in the original graph (RedditM). 
    \textbf{Right}: similarity distribution of edges added by the cosine refiner compared to disconnected pairs. 
    }
    \label{fig:refinement}
\end{wrapfigure}

\subsubsection{Effects of Graph Refinement}
\label{sec:prelim}
\vspace{-2mm}
To validate the effect of topology refinement, we compare the semantic similarity distribution of edges before and after refinement. 
As shown in Fig.~\ref{fig:refinement}, the original graph exhibits substantial overlap between connected and disconnected node pairs, indicating noisy and missing edges. 
In contrast, the edges introduced by the refiner are concentrated in the high-similarity region, with significantly reduced overlap. 
This demonstrates that the refiner selectively recovers semantically meaningful connections while avoiding spurious links, leading to a cleaner and more informative graph structure.

This shift in distribution provides direct evidence that ReCoG improves topology quality by increasing semantic consistency of graph edges. More detailed analysis is provided in Appendix~\ref{app:misalign}.

\subsection{Semantic Alignment via Cross-Modal Message Passing}
\label{sec:gnn}
\vspace{-2mm}
Given the refined graph $\mathbf{A}_{\text{refine}}$, we propagate information jointly across modalities and graph structure. We first present the propagation rule directly through its supergraph formulation, which exposes its spectral and alignment properties (Sec.~\ref{sec:theory}), and then show that it admits an efficient per-modality implementation.

We first project modality-specific features into a common latent space of dimension $d_h$:
\begin{align}
\mathbf{h}_{i}^{m, (0)} &= \text{Proj}(\mathbf{x}_i^{m}; \phi^{m}), \quad m \in \{t, v\},
\end{align}
where $\phi^{m}$ parametrizes the projection (a linear layer followed by layer normalization (LN) and the GELU nonlinearity with dropout). This ensures dimensional alignment and stabilizes subsequent cross-modal interaction. We then adopt symmetric normalization of $\mathbf{A}_{\text{refine}}$ with self-loops.

\textbf{Coupled propagation on a modality-augmented supergraph.}
We define the stacked representation $\mathbf{H}^{(\ell)} = [\mathbf{H}^{t,(\ell)};\, \mathbf{H}^{v,(\ell)}]$ and the supergraph adjacency
\begin{equation}
\mathbf{A}_{\text{sup}} =
\begin{bmatrix}
\mathbf{A}_{\text{refine}} & \gamma \mathbf{I} + \rho \mathbf{A}_{\text{refine}} \\
\gamma \mathbf{I} + \rho \mathbf{A}_{\text{refine}} & \mathbf{A}_{\text{refine}}
\end{bmatrix},
\label{eq:A-sup}
\end{equation}
where $\gamma, \rho \in (0,1)$ are learnable coupling scalars (parameterized via sigmoid). Propagation then takes the standard form
\begin{equation}
\mathbf{H}^{(\ell+1)} = \sigma\big(\mathbf{A}_{\text{sup}} \mathbf{H}^{(\ell)} \mathbf{W}^{(\ell)}\big),
\label{eq:supergraph-update}
\end{equation}
with a \emph{single shared} weight matrix $\mathbf{W}^{(\ell)}$ acting on both modality blocks. Two design choices are central:
(1) \textbf{The symmetric block construction}(identical diagonals and off-diagonals) 
places the two modalities on equal footing and enables decomposition into 
symmetric and anti-symmetric eigenmodes (Corollary~\ref{cor:spectral}). 
(2) \textbf{Sharing $\mathbf{W}^{(\ell)}$ across modalities} is what enables the 
cross-modal contraction guarantee in Theorem~\ref{prop:alignment}; 
a separate-weight variant would introduce a residual term proportional to 
$\|\mathbf{W}_T^{(\ell)} - \mathbf{W}_V^{(\ell)}\|$ that does not vanish 
even when the two modalities have aligned.

\textbf{Per-modality view.}
Expanding the block product in Eq.~\eqref{eq:supergraph-update} recovers an explicit per-modality update. For the textual stream:
\begin{align}
\mathbf{h}_{i}^{t, (\ell+1)} = \sigma\Bigg(\mathbf{W}^{(\ell)} \cdot \Big(
        \underbrace{\sum_{j \in \mathcal{N}(i)} A_{ij}\, \mathbf{h}_{j}^{t,(\ell)}}_{\text{intra-modal aggregation}}
        \;+\; \underbrace{\gamma \cdot \mathbf{h}_{i}^{v, (\ell)}}_{\text{cross-modal self-injection}}
        \;+\; \underbrace{\rho \cdot \sum_{j \in \mathcal{N}(i)} A_{ij}\, \mathbf{h}_{j}^{v, (\ell)}}_{\text{cross-modal neighbor aggregation}}
    \Big)\Bigg),
\label{eq:cross_modal}
\end{align}
where $A_{ij} = \mathbf{A}_{\text{refine}}(i,j)$ (Eqn.~\eqref{eq:adj_sim} or~\eqref{eq:adj_res}) and $\mathcal{N}(i)$ is the top-$k$ semantic neighbors (Eqn.~\eqref{eq:sim_topk} or Eqn.~\eqref{eqn:res_topk}). The visual update $\mathbf{h}_{i}^{v, (\ell+1)}$ is defined symmetrically. Each term has a distinct role: intra-modal aggregation propagates within a stream, self-injection mixes the node's own counterpart-modality representation, and cross-modal neighbor aggregation propagates counterpart-modality signals over the refined graph. Importantly, this formulation reduces the number of nodes by half and decreases the number of edges from $4E+2N$ in Eq.~\eqref{eq:supergraph-update} to $2E$, making it more efficient.

\textbf{Complexity optimization.}
The per-modality view above avoids materializing 
$\mathbf{A}_{\text{sup}}$, but still requires four GNN calls per layer 
(two per modality: one for intra-modal and one for cross-modal neighbor 
aggregation). By exploiting linearity of aggregation,
\begin{equation}
\sum_j A_{ij}(\mathbf{h}_{j}^t + \rho \mathbf{h}_{j}^v) = \sum_j A_{ij}\mathbf{h}_{j}^t + \rho \sum_j A_{ij}\mathbf{h}_{j}^v,
\end{equation}
we merge intra-modal and cross-modal neighbor aggregation into a single GNN call per modality:
\begin{align}
\tilde{\mathbf{h}}_i^{t,(\ell)} = \text{GNN}(\mathbf{h}_j^{t,(\ell)} + \rho\, \mathbf{h}_j^{v,(\ell)},\, \mathbf{A}_{\text{refine}}), \qquad
\tilde{\mathbf{h}}_i^{v,(\ell)} = \text{GNN}(\mathbf{h}_j^{v,(\ell)} + \rho\, \mathbf{h}_j^{t,(\ell)},\, \mathbf{A}_{\text{refine}}),
\end{align}
followed by the self-injection term:
\begin{align}
\mathbf{h}_i^{t,(\ell+1)} = \sigma\!\Big(\mathrm{LN}\!\big(\tilde{\mathbf{h}}_i^{t,(\ell)} + \gamma\, \mathbf{W}^{(\ell)} \mathbf{h}_i^{v,(\ell)}\big)\Big), \quad \mathbf{h}_i^{v,(\ell+1)} = \sigma\!\Big(\mathrm{LN}\!\big(\tilde{\mathbf{h}}_i^{v,(\ell)} + \gamma\, \mathbf{W}^{(\ell)} \mathbf{h}_i^{t,(\ell)}\big)\Big).
\end{align}
This reduces the cost from four to two calls per layer and is exactly equivalent to Eq.~\eqref{eq:supergraph-update} for any 
linear aggregation operator. A detailed complexity optimization analysis is provided in Appendix~\ref{app:opt}.

\textbf{Closed loop with the refiner.}
Combined with the graph refiner of Sec.~\ref{sec:refiner}, the propagation step closes a loop: at each training step, refined representations sharpen the topology produced by Eq.~\eqref{eq:adj_sim}/\eqref{eq:adj_res}, and the updated topology in turn shapes the next round of cross-modal propagation. This is the reciprocal co-evolution formalized in Proposition~\ref{prop:joint}: fixed refiners (i.e., not jointly optimized with the GNN) are recovered as special cases, while ReCoG additionally allows task gradients to update the topology refiner, strictly enlarging the feasible optimization space.

\begin{proposition}[\textbf{Joint Optimization Contains Non-Reciprocal Alternatives}]
\label{prop:joint}
Let 
$\mathcal{J}(\theta_{\mathrm{ref}},\theta_{\mathrm{gnn}})
=
\ell(f_{\theta_{\mathrm{gnn}}}(X,A_{\mathrm{refine}}(\theta_{\mathrm{ref}})),Y)$
denote the end-to-end task objective. Then the joint optimization space contains any fixed or frozen refiner as a special case:
\begin{equation}
    \inf_{\theta_{\mathrm{ref}},\,\theta_{\mathrm{gnn}}}
    \mathcal{J}(\theta_{\mathrm{ref}},\theta_{\mathrm{gnn}})
    \le
    \inf_{\theta_{\mathrm{gnn}}}
    \mathcal{J}(\theta_{\mathrm{ref}}^{\mathrm{fix}},\theta_{\mathrm{gnn}}),
    \label{eq:joint_dominance}
\end{equation}
for any fixed $\theta_{\mathrm{ref}}^{\mathrm{fix}}$, including the identity refiner
$A_{\mathrm{refine}}=A_{\mathrm{orig}}$ when it belongs to the refiner parameter space, and any frozen two-stage refiner.
\end{proposition}

\subsection{Adaptive Modality Fusion}
\label{sec:fusion}
\vspace{-2mm}
After $L$ layers of coupled message passing, we obtain modality-specific node representations
$\mathbf{z}_i^t = \mathbf{h}_i^{t,(L)}$ and $\mathbf{z}_i^v = \mathbf{h}_i^{v,(L)}$.
While the two modalities have been progressively aligned through shared propagation and cross-modal interaction, they may still carry complementary or unevenly informative signals. We therefore fuse them into a unified representation through a \emph{node-adaptive gating mechanism}. Specifically, we compute a node-specific scalar gate:
\begin{equation}
    g_i = \sigma\Big(\text{MLP}\big([\mathbf{z}_i^t \, ; \, \mathbf{z}_i^v]\big)\Big) \in (0,1),
    \label{eq:gate-rewrite}
\end{equation}
where $\sigma(\cdot)$ is the sigmoid function and $[\cdot ; \cdot]$ denotes concatenation.
The final representation is:
\begin{equation}
    \mathbf{z}_i = \text{LN}\Big(
    g_i \cdot \mathbf{z}_i^t + (1 - g_i) \cdot \mathbf{z}_i^v
    \Big).
    \label{eq:fuse-rewrite}
\end{equation}

\textbf{Relation to coupled propagation.}
Unlike early or late fusion strategies, this design separates two roles:
(i) \textit{Cross-modal message passing} aligns and propagates multimodal information across the graph, and 
(ii) \textit{Adaptive fusion} performs final integration based on node-specific relevance.

This separation allows ReCoG to first establish structurally grounded cross-modal representations and then perform flexible, task-adaptive combination. 
This final adaptive fusion ensures that the benefits of reciprocal co-evolution are translated into task-specific representations.

\section{Theoretical Analysis}
\label{sec:theory}
\vspace{-2mm}
We provide formal justifications for the design choices in ReCoG from two complementary perspectives: the representational benefit of coupled cross-modal propagation, and the cross-modal alignment guarantee of shared-weight design. All proofs including propositions are deferred to Appendix~\ref{app:proof_1}-~\ref{app:proof_4}.

\begin{theorem}[\textbf{Layerwise Cross-Modal Information Flow}]
\label{prop:coupled}
Define a multimodal GNN as \emph{decoupled} if each modality stream's update at layer $\ell$ depends only on that stream's representations from layer $\ell - 1$ and the graph $A$, with arbitrary per-modality weight matrices $W_T^{(\ell)}, W_V^{(\ell)}$.  Then the cross-modal Jacobian satisfies:
\begin{equation}
    \frac{\partial\, H_T^{(\ell)}}{\partial\, H_V^{(0)}}
    \;=\;
    \begin{cases}
        0 & \text{for any decoupled model},\; \forall\, \ell \ge 0, \\[4pt]
        \neq 0 \;\text{(generically)} & \text{for the coupled model with } \rho > 0 \text{ or } \gamma > 0,\; \forall\, \ell \ge 1.
    \end{cases}
    \label{eq:cross_modal_jacobian}
\end{equation}
Coupled propagation enables layerwise cross-modal information flow that is structurally impossible in any decoupled architecture, regardless of per-modality parameterization or model capacity. This distinguishes per-layer cross-modal coupling from late fusion (e.g., MMGCN), which cannot refine one modality using the graph neighbors of the other during propagation.
\end{theorem}

The supergraph interpretation (Eq.~\eqref{eq:A-sup}) enables an explicit spectral characterization of how the coupling coefficients reshape the effective frequency response.
 
\begin{corollary}[\textbf{Spectral Characterization of Coupled Propagation}]
\label{cor:spectral}
Let $\mu_1 \ge \mu_2 \ge \cdots \ge \mu_N$ be the eigenvalues of $A_{\mathrm{refine}}$. Then the $2N$ eigenvalues of $A_{\mathrm{sup}}$ are
\begin{equation}
    \lambda_i^{+} = (1+\rho)\,\mu_i + \gamma,
    \qquad
    \lambda_i^{-} = (1-\rho)\,\mu_i - \gamma,
    \qquad i=1,\dots,N,
    \label{eq:spectral}
\end{equation}
where $\lambda_i^{+}$ correspond to eigenmodes in which text and vision components are aligned, and $\lambda_i^{-}$ to eigenmodes in which they oppose.
Intuitively, $\rho$ and $\gamma$ act as a learned spectral filter that amplify symmetric modes and attenuate anti-symmetric modes.
\end{corollary}

\begin{theorem}[\textbf{Shared-Weight Cross-Modal Consistency}]
\label{prop:alignment}
Consider one coupled layer with shared weight $W^{(\ell)}$ and $1$-Lipschitz activation $\sigma(\cdot)$. Define the cross-modal discrepancy $\Delta^{(\ell)} = H_T^{(\ell)} - H_V^{(\ell)}$. Then
\begin{equation}
    \bigl\|\Delta^{(\ell+1)}\bigr\|
    \;\le\;
    \bigl\|W^{(\ell)}\bigr\|
    \Bigl(
        (1-\rho)\,\bigl\|A_{\mathrm{refine}}\bigr\| + \gamma
    \Bigr)
    \bigl\|\Delta^{(\ell)}\bigr\|.
    \label{eq:contraction}
\end{equation}
Consequently, after $L$ layers:
\begin{equation}
    \bigl\|\Delta^{(L)}\bigr\|
    \;\le\;
    \prod_{\ell=0}^{L-1}
    \bigl\|W^{(\ell)}\bigr\|
    \Bigl((1-\rho)\,\bigl\|A_{\mathrm{refine}}\bigr\| + \gamma\Bigr)
    \;\cdot\;
    \bigl\|\Delta^{(0)}\bigr\|.
    \label{eq:contraction_L}
\end{equation}
When the per-layer factor $\|W^{(\ell)}\|\bigl((1-\rho)\|A_{\mathrm{refine}}\| + \gamma\bigr) < 1$, the discrepancy contracts exponentially with depth, ensuring that text and vision embeddings converge to a shared representation space. 
\end{theorem}

\vspace{-2mm}
\section{Experiments}
\label{sec:exp}
\vspace{-2mm}

\textbf{Datasets.}
We evaluate ReCoG on two tasks: node classification (NC) and link prediction (LP) across \textbf{nine} multimodal graph datasets spanning diverse domains. 
For NC, we consider two product-review graphs, \textbf{Ele-Fashion} and \textbf{Goodreads-NC}~\citep{zhu2025mosaic}, and five datasets from the MAGB benchmark~\citep{yan2025graph}, including three e-commerce graphs (\textbf{Movies}, \textbf{Toys}, \textbf{Grocery}) and two social-interaction graphs (\textbf{Reddit-S}, \textbf{Reddit-M}). 
For LP, we use two Amazon co-purchase graphs, \textbf{Amazon-Sports} and \textbf{Amazon-Cloth}~\citep{zhu2025mosaic}. 
Each node is associated with a text description and an image, from which pretrained encoders extract modality-specific feature vectors. 
Detailed dataset statistics are provided in Table~\ref{tab:dataset-stats} in the Appendix. All experiments were conducted on a single NVIDIA A100 GPU.

\textbf{Encoders.}
To ensure fair and comprehensive evaluation, we follow benchmark protocols and consider multiple encoder configurations. 
For Ele-Fashion, Goodreads-NC, and LP datasets, we use four combinations: \textit{CLIP}, \textit{T5+ViT}, \textit{ImageBind}, and \textit{T5+DINOv2}. 
CLIP and ImageBind produce aligned multimodal embeddings, while ViT/DINOv2 serve as vision encoders and T5 as the text encoder.

For MAGB datasets, we adopt three configurations: \textit{L+C} (LLaMA-3.1-8B + CLIP), \textit{QWen} (QWen2-VL-7B), and \textit{LLaVL} (LLaMA-3.2-11B-Vision). 
A key distinction is that QWen and LLaVL baselines directly use \emph{jointly pretrained multimodal embeddings}, where text and image representations are inherently aligned. 
In contrast, ReCoG treats text and image features as \emph{separate streams} and learns cross-modal alignment during training. 
As shown in our results, this learned alignment consistently matches or surpasses encoder-level alignment, highlighting the effectiveness of our architecture.

\begin{table*}[t]
\centering
\caption{Node classification accuracy (\%) on the  MM-Graph~\citep{zhu2025mosaic}. \textcolor{orange}{Best} and \textcolor{teal}{runner-up} are highlighted.}
\scriptsize
\setlength{\tabcolsep}{3.5pt}
\renewcommand{\arraystretch}{1.15}
\begin{tabular}{l l *{4}{c} *{4}{c}}
\toprule
& & \multicolumn{4}{c}{\textbf{Ele-fashion}} & \multicolumn{4}{c}{\textbf{Goodreads-NC}} \\
\cmidrule(lr){3-6} \cmidrule(lr){7-10}
& \textbf{Model} & CLIP & T5+ViT & IBind & T5+DINO & CLIP & T5+ViT & IBind & T5+DINO \\
\midrule
& MMGCN     & 86.10\,$\pm$\,0.50 & 82.39\,$\pm$\,0.30 & 86.21\,$\pm$\,0.94 & 85.53\,$\pm$\,0.33 & 83.29\,$\pm$\,0.20 & 81.85\,$\pm$\,0.22 & 80.58\,$\pm$\,1.08 & 82.44\,$\pm$\,0.11 \\
& MGAT      & 84.66\,$\pm$\,0.29 & 84.01\,$\pm$\,0.08 & 86.12\,$\pm$\,0.08 & 84.54\,$\pm$\,0.27 & 76.48\,$\pm$\,0.59 & 75.43\,$\pm$\,0.76 & 69.45\,$\pm$\,6.25 & 74.98\,$\pm$\,1.23 \\
& GCN       & 79.83\,$\pm$\,0.03 & 79.63\,$\pm$\,0.07 & 80.35\,$\pm$\,0.02 & 79.37\,$\pm$\,0.04 & 81.61\,$\pm$\,0.01 & 81.67\,$\pm$\,0.03 & 78.91\,$\pm$\,0.04 & 81.71\,$\pm$\,0.03 \\
& SAGE      & 87.10\,$\pm$\,0.02 & 84.41\,$\pm$\,0.09 & 87.71\,$\pm$\,0.13 & 85.31\,$\pm$\,0.09 & 83.30\,$\pm$\,0.02 & 83.03\,$\pm$\,0.04 & 80.39\,$\pm$\,0.21 & 82.99\,$\pm$\,0.08 \\
& MLP       & 85.16\,$\pm$\,0.03 & 84.98\,$\pm$\,0.05 & 88.73\,$\pm$\,0.01 & 84.87\,$\pm$\,0.01 & 72.29\,$\pm$\,0.02 & 67.82\,$\pm$\,0.07 & 58.75\,$\pm$\,0.05 & 68.83\,$\pm$\,0.03 \\
\midrule
\multirow{2}{*}{\rotatebox{90}{\textit{sim}}}
& ReCoG-GCN  & 88.56\,$\pm$\,0.05 & 87.25\,$\pm$\,0.08 & 89.01\,$\pm$\,0.03 & 87.70\,$\pm$\,0.06 & 90.43\,$\pm$\,0.09 & 88.18\,$\pm$\,0.10 & 89.26\,$\pm$\,0.08 & 89.21\,$\pm$\,0.12 \\
& ReCoG-SAGE & \cellcolor{orange!50}89.01\,$\pm$\,0.05 & \cellcolor{teal!30}88.02\,$\pm$\,0.07 & \cellcolor{teal!30}89.31\,$\pm$\,0.10 & 88.21\,$\pm$\,0.10 & 90.51\,$\pm$\,0.08 & 88.24\,$\pm$\,0.11 & 89.31\,$\pm$\,0.09 & 89.30\,$\pm$\,0.10 \\
\midrule
\multirow{2}{*}{\rotatebox{90}{\textit{res}}}
& ReCoG-GCN  & 88.60\,$\pm$\,0.08 & 87.99\,$\pm$\,0.09 & 89.06\,$\pm$\,0.10 & \cellcolor{teal!30}88.28\,$\pm$\,0.08 & \cellcolor{orange!50}90.68\,$\pm$\,0.06 & \cellcolor{orange!50}90.98\,$\pm$\,0.03 & \cellcolor{teal!30}90.05\,$\pm$\,0.03 & \cellcolor{orange!50}90.19\,$\pm$\,0.06 \\
& ReCoG-SAGE & \cellcolor{teal!30}88.92\,$\pm$\,0.07 & \cellcolor{orange!50}88.04\,$\pm$\,0.06 & \cellcolor{orange!50}89.51\,$\pm$\,0.03 & \cellcolor{orange!50}88.34\,$\pm$\,0.06 & \cellcolor{teal!30}90.66\,$\pm$\,0.11 & \cellcolor{teal!30}90.07\,$\pm$\,0.11 & \cellcolor{orange!50}90.25\,$\pm$\,0.03 & \cellcolor{teal!30}90.08\,$\pm$\,0.05 \\
\bottomrule
\end{tabular}
\label{tab:nc-main}
\end{table*}
\begin{table*}
\centering
\caption{Node classification accuracy (\%) on the MAGB datasets~\citep{yan2025graph}. 
}
\scriptsize
\setlength{\tabcolsep}{3pt}
\renewcommand{\arraystretch}{1.15}
\begin{threeparttable}
\resizebox{\textwidth}{!}{%
\begin{tabular}{l l *{3}{c} *{3}{c} *{3}{c} *{3}{c} *{3}{c}}
\toprule
& & \multicolumn{3}{c}{\textbf{Movies}} & \multicolumn{3}{c}{\textbf{Toys}} & \multicolumn{3}{c}{\textbf{Grocery}} & \multicolumn{3}{c}{\textbf{Reddit-S}} & \multicolumn{3}{c}{\textbf{Reddit-M}} \\
\cmidrule(lr){3-5} \cmidrule(lr){6-8} \cmidrule(lr){9-11} \cmidrule(lr){12-14} \cmidrule(lr){15-17}
& \textbf{Model} & L+C & QWen & LLaVL & L+C & QWen & LLaVL & L+C & QWen & LLaVL & L+C & QWen & LLaVL & L+C & QWen & LLaVL \\
\midrule
& GCN  & 53.98 & 55.19 & 54.92 & 81.59 & 80.95 & 81.57 & 84.64 & 82.96 & 84.17 & 89.77 & 90.88 & 91.13 & 71.42 & 72.96 & 75.64 \\
& SAGE & 56.08 & 58.01 & 57.39 & 82.35 & 81.45 & 82.35 & 86.65 & 85.55 & 86.81 & 91.08 & 92.57 & 92.94 & 80.84 & 81.23 & 85.61 \\
\midrule
\multirow{2}{*}{\rotatebox{90}{\textit{sim}}}
& ReCoG-GCN  & 56.97 & 57.72 & 57.51 & 81.98 & 81.20 & 82.19 & 85.89 & 83.46 & 85.03 & \cellcolor{orange!50}97.20 & \cellcolor{orange!50}96.98 & \cellcolor{orange!50}97.26 & \cellcolor{orange!50}90.55 & \cellcolor{orange!50}90.24 & \cellcolor{orange!50}90.87 \\
& ReCoG-SAGE & 56.94 & 58.92 & \cellcolor{teal!30}60.00 & \cellcolor{teal!30}82.58 & \cellcolor{teal!30}82.27 & \cellcolor{teal!30}82.75 & \cellcolor{teal!30}86.86 & 86.24 & \cellcolor{orange!50}89.05 & 96.76 & 96.38 & 96.73 & 88.75 & 89.18 & \cellcolor{teal!30}90.11 \\
\midrule
\multirow{2}{*}{\rotatebox{90}{\textit{res}}}
& ReCoG-GCN  & \cellcolor{teal!30}57.45 & \cellcolor{teal!30}60.12 & 58.62 & 82.07 & \cellcolor{teal!30}82.27 & 82.27 & 86.18 & \cellcolor{teal!30}86.39 & 86.56 & \cellcolor{teal!30}96.89 & \cellcolor{teal!30}96.82 & 96.73 & 87.28 & 87.31 & 87.42 \\
& ReCoG-SAGE & \cellcolor{orange!50}57.93 & \cellcolor{orange!50}60.84 & \cellcolor{orange!50}60.45 & \cellcolor{orange!50}82.68 & \cellcolor{orange!50}82.36 & \cellcolor{orange!50}83.18 & \cellcolor{orange!50}86.92 & \cellcolor{orange!50}86.56 & \cellcolor{teal!30}87.53 & \cellcolor{teal!30}96.89 & 96.48 & \cellcolor{teal!30}97.17 & \cellcolor{teal!30}89.18 & \cellcolor{teal!30}89.71 & 89.86 \\
\bottomrule
\end{tabular}}
\begin{tablenotes}
\footnotesize
\item[$\dagger$] Following the MAGB benchmark protocol~\cite{yan2025graph}, we report the average 
over 10 random seeds for node classification.
\end{tablenotes}
\end{threeparttable}
\label{tab:magb}
\end{table*}

\begin{table}
\centering
\caption{Comparison with recent multimodal graph methods. R-S = Reddit-S, R-M = Reddit-M.}
\label{tab:compare-sota}
\small
\setlength{\tabcolsep}{4pt}
\renewcommand{\arraystretch}{1.2}
\resizebox{0.85\textwidth}{!}{
\begin{threeparttable}
\begin{tabular}{l ccccccc cc}
\toprule
& \multicolumn{7}{c}{\textbf{Node Classification (Acc. \%)}} & \multicolumn{2}{c}{\textbf{Link Pred. (MRR \%)}} \\
\cmidrule(lr){2-8} \cmidrule(lr){9-10}
\textbf{Model} & Fashion & GoodReads & Movies & Toys & Grocery & R-S & R-M & Sports & Cloth \\
\midrule
NSG-MoE    & 88.42 & --- & 53.98 & 80.92 & 86.28 &95.93 & 85.03 & 34.39 & 24.10 \\
LION       & --- & 78.54 & 58.61 & --- & 88.30 & 94.80 & --- & --- & --- \\
PLANET     & 87.37 & 84.16 & 57.06 & 81.22 & 85.16 & 96.62 & --- & 27.51 & 20.25 \\
UniGraph2  & 87.91 & 81.97 & *52.92 & *79.49 & *85.24 & *93.50 & *70.19 & 31.61 & 25.01 \\
\midrule
\textbf{ReCoG-sim} & \cellcolor{teal!30}89.31 & \cellcolor{teal!30}90.51 & \cellcolor{teal!30}60.00 & \cellcolor{teal!30}82.75 & \cellcolor{orange!50}89.05 & \cellcolor{orange!50}97.26 & \cellcolor{orange!50}90.87 & \cellcolor{teal!30}38.53 & \cellcolor{teal!30}28.30 \\
\textbf{ReCoG-res} & \cellcolor{orange!50}89.51 & \cellcolor{orange!50}90.98 & \cellcolor{orange!50}60.84 & \cellcolor{orange!50}83.18 & \cellcolor{teal!30}87.53 & \cellcolor{teal!30}97.17 & \cellcolor{teal!30}89.86 & \cellcolor{orange!50}38.57 & \cellcolor{orange!50}28.73 \\
\bottomrule
\end{tabular}
\begin{tablenotes}
\footnotesize
\item[*] Results were not reported in the original paper, we use the numbers from ~\citep{zhang2026modality,li2026lion}.
\item ``---'' indicates the method was not evaluated on that dataset.
\end{tablenotes}
\end{threeparttable}
}
\end{table}

\textbf{Baselines.} Our baselines fall into two groups: (i) multimodal GNNs that operate on fixed graphs (MMGCN and MGAT~\citep{zhu2025mosaic} with late fusion, GCN~\citep{kipf2016semi}/SAGE~\citep{hamilton2017inductive}/MLP on concatenated features), and (ii) recent multimodal graph models, including \textit{NSG-MoE}~\citep{zhang2026modality}, \textit{LION}~\citep{li2026lion}, \textit{PLANET}~\citep{liu2026toward}, and \textit{UniGraph2}~\citep{he2025unigraph2}. Note that, following the settings of major baselines and benchmarks~\citep{zhu2025mosaic,yan2025graph}, we focus on multimodal-aware baselines, as most structure learning methods are designed to enhance single-modality graphs without semantic consideration and would require nontrivial adaptation to handle cross-modal evidence fairly.

\textbf{Evaluation.}
We evaluate two variants of our model: \textbf{ReCoG-sim} (similarity refiner) and \textbf{ReCoG-res} (residual refiner), both using the same cross-modal message passing backbone. 
Each variant is tested with both GCN-style and SAGE-style convolutions. 
We follow the same data splits and experimental settings as prior work, including multiple runs with different random seeds.
For fair comparison, we reproduce results for \textit{MMGCN}, \textit{MMGAT}, and \textit{MLP}, while reporting official results for recent state-of-the-art methods. 
For models with substantially different architectures (e.g., \textit{UniGraph2}), we use the best reported results from their original papers.

\subsection{Node Classification Results}
\label{sec:nc_results}
\vspace{-2mm}

Tables~\ref{tab:nc-main} and~\ref{tab:magb}  report node classification accuracy across all datasets and encoder configurations. Table~\ref{tab:compare-sota} compares the best performance between our ReCoG and SOTA multimodal graph models. 

\textbf{Overall performance.}
ReCoG consistently outperforms all baselines across datasets, encoders, and backbone architectures. 
The improvements are particularly significant on large and structurally complex graphs, indicating that jointly refining topology and representations is especially beneficial when the observed graph is noisy or incomplete.

\textbf{Key observations.}
(i) \textbf{Strong gains on large scale datasets.}
On Goodreads-NC, ReCoG-res achieves $90.98\%$, outperforming the strongest recent baseline (PLANET, $84.16\%$) by a large margin.
On Reddit-M, ReCoG improves over GCN by more than $15\%$, demonstrating substantial robustness. We attribute this to the fact that large-scale interaction graphs tend to contain a higher proportion of semantically irrelevant edges, and the refiner mitigates this issue by down-weighting noisy edges and introducing semantically coherent ones.
(ii) \textbf{Consistent improvements across encoders.}
ReCoG improves performance under all encoder configurations, including CLIP, ImageBind, and VLM-based encoders (QWen, LLaVL), indicating that the gains are not tied to a specific feature extractor.
(iii) \textbf{Graph-level alignment surpasses encoder-level alignment.}
Even when using VLM embeddings that are already aligned (e.g., QWen, LLaVL), ReCoG still achieves significant gains (e.g., $+6.10\%$ on Reddit-S and $+17.28\%$ on Reddit-M).

\textbf{Remark.}
These results highlight two key advantages of ReCoG.
First, the graph refiner improves topology by introducing semantically meaningful connections and removing noisy edges.
Second, cross-modal message passing enables information exchange during propagation rather than relying on static encoder alignment.
Together, these components produce representations that are better aligned with both graph structure and multimodal semantics.

\textbf{Effect of refiner variants.}
The residual refiner generally achieves the best performance, suggesting that its expressive edge scoring better captures complex multimodal relationships.
However, the similarity refiner remains competitive and sometimes preferable when pretrained embedding geometry is already informative, due to its strong inductive bias and lower risk of overfitting.

\subsection{Link Prediction Results}
\label{sec:lp-results}
\vspace{-2mm}
Table~\ref{tab:lp-main} (in the Appendix due to space limit) reports link prediction performance on Amazon-Sports and Amazon-Cloth, evaluated using MRR, Hits@1, and Hits@10.
ReCoG consistently outperforms all baselines across both datasets and all encoder configurations, demonstrating robust improvements in LP quality.

\textbf{Key observations.}
\textit{First}, ReCoG achieves substantial gains in MRR. On Amazon-Sports with CLIP features, ReCoG-res (GCN) reaches 38.57\% MRR, compared to 33.83\% from the strongest baseline (SAGE). On Amazon-Cloth, ReCoG-res (GCN) achieves 28.73\% MRR, outperforming SAGE (25.20\%) by nearly 3 points. 
\textit{Second}, improvements are particularly pronounced in Hits@10. For example, on Amazon-Sports with ImageBind features, ReCoG-res (GCN) achieves 75.60\% H@10 versus 65.61\% for GCN. This suggests that the refined topology is especially effective at expanding the set of high-quality candidate links.

\textbf{Variant comparison.}
The two refiner variants show similar trends to node classification but with smaller differences. ReCoG-sim performs slightly better on Amazon-Sports across most encoders, while ReCoG-res shows marginal advantages on Amazon-Cloth. This reduced gap indicates that, in link prediction, performance is more sensitive to the quality of the learned representations than to the specific refinement mechanism, since the decoder directly operates on embedding similarities.
\vspace{-3mm}

\subsection{Ablation Studies}
\label{sec:ablation}
\vspace{-2mm}

\textbf{Synergistic Effect.} We analyze the contribution of each component in ReCoG by comparing the full model with variants that remove (i) the graph refiner, (ii) cross-modal message passing, or (iii) both. 
The variant without both components reduces to independent modality-specific GNNs with late fusion. 
Experiments are conducted on three datasets using GCN as the backbone. 
We use CLIP features for Ele-Fashion, and CLIP+LLaMA for Movies and Reddit-S. 
Results are reported in Table~\ref{tab:component}.

\textit{Overall effect of components.}
Removing either component consistently degrades performance, while removing both leads to the largest drop. 
This confirms that both topology refinement and cross-modal interaction are essential, and their combination provides the strongest gains.

\begin{wraptable}{r}{0.6\textwidth}
\centering
\caption{Ablation study on node classification accuracy (\%). $\dagger$ \textit{MMGCN} uses separate GCNs for each modality with late fusion.}
\label{tab:component}
\resizebox{1.0\linewidth}{!}{
\begin{tabular}{cl|ccc}
\toprule
& \textbf{Configuration} &\textbf{Fashion (Clip)} &\textbf{Movies (L+C)} & \textbf{Reddit-S (L+C)} \\
\midrule
& MMGCN$^\dagger$ (w/o both) & 86.10 & 54.18 & 91.13  \\
\midrule
\multirow{3}{*}{\rotatebox[origin=c]{90}{sim}}
 & w/o Cross-Modal & 86.83 & 55.16 & 94.28  \\
 & w/o Refiner      & 87.09 &  56.19  & 95.97   \\
 & Full Model    & 88.56 & 56.97 & 97.20   \\
\midrule
\multirow{3}{*}{\rotatebox[origin=c]{90}{res}}
 & w/o Cross-Modal & 86.41 & 55.74 & 92.74   \\
 & w/o Refiner      & 87.09 &  56.19  & 95.97   \\
 & Full Model   & 88.60 & 57.45  & 96.89   \\
\bottomrule
\end{tabular}
}
\end{wraptable}

\textit{Synergistic interaction.}
A key observation is that the benefit of the refiner depends on whether cross-modal message passing is present. 
When cross-modal coupling is removed, the similarity refiner performs comparably to, or slightly better than, the residual refiner. 
However, once cross-modal interaction is enabled, the residual refiner consistently achieves the best performance.

\textit{Analysis.}
This behavior reflects the interaction between topology learning and representation learning.
The residual refiner constructs a \emph{joint multimodal topology} based on pairwise features $\varphi(i,j)$ that combine text and vision signals. 
When the two modality streams are processed independently, this unified topology becomes suboptimal, as the optimal neighborhood structure may differ across modalities. 
In addition, the absence of cross-modal interaction weakens the supervision signal for learning expressive edge scoring functions.
In contrast, the similarity refiner relies on cosine similarity with a learnable modality weight, which provides a strong inductive bias. 
When one modality dominates, this bias allows it to produce a robust topology even without cross-modal interaction.
When cross-modal coupling is introduced, the downstream GNN aggregates mixed-modality information at each layer. 
In this setting, a jointly learned topology becomes better aligned with the propagation mechanism, and the residual refiner can fully exploit its higher expressive power. 
The enriched cross-modal gradients further improve its learning dynamics, enabling it to surpass the similarity-based variant.

\textit{Takeaway.}
These results highlight that topology refinement and cross-modal propagation are \emph{not independent}: 
the effectiveness of a more expressive refiner is unlocked only when coupled with cross-modal message passing, supporting the design of reciprocal co-evolution in ReCoG.

\textbf{Top-K Selection (Eqns.~\eqref{eq:sim_topk} ,~\eqref{eqn:res_topk}).}
Table~\ref{tab:topk} examines the sensitivity of both refiners to the number of candidate $K$. The results show that our model's performance changes slightly within a reasonable range of $K$, demonstrating the stability of our design. A detailed analysis is provided in Appendix~\ref{app:topk}.

\section{Conclusion}
\label{sec:conclu}
\vspace{-2mm}
We propose ReCoG, a co-evolution framework that jointly refines graph topology and learns cross-modal node representations end-to-end for multimodal attributed graphs.
Unlike prior methods that treat topology as fixed or perform modality fusion only at the final stage, ReCoG enables the refined structure and learned embeddings to mutually reinforce each other throughout training, via a configurable Multimodal Graph Refiner and a Cross Modal Message Passing.
Extensive experiments on nine datasets spanning node classification and link prediction demonstrate consistent improvements over both classical and recent state-of-the-art baselines. Moreover, our approach is parallelizable to standard GNN training pipelines and shows that learned cross-modal coupling can complement, and in some cases substitute for, the alignment quality of jointly pretrained VLMs.

\bibliographystyle{ACM-Reference-Format}
\bibliography{ref2026}

\newpage
\appendix

\section{Broader Impacts}
\label{sec:impact}

Our proposed ReCoG advances multimodal graph learning by enabling joint refinement of graph structure and cross-modal representations, which can improve performance in applications such as recommendation systems, social networks, and scientific discovery. By leveraging multimodal semantic signals to infer graph connectivity, the proposed approach can better capture meaningful relationships in complex data.

However, these capabilities also introduce risks. Since ReCoG actively modifies graph topology, it may introduce spurious or biased connections when multimodal features are noisy or misaligned, potentially amplifying existing biases in pretrained encoders or datasets. In applications involving user data, such as social or e-commerce graphs, this may lead to unfair recommendations or unintended profiling.

Additionally, the ability to infer latent relationships from multimodal signals raises privacy concerns, as the model may uncover sensitive or hidden connections between entities. Misuse in surveillance or user behavior modeling contexts is also possible.

Future work should explore bias mitigation, robustness to noisy modalities, and privacy-preserving mechanisms to ensure responsible deployment.

\section{Related Work}
\label{app:related}

\textbf{Graph Neural Networks.}
Graph Neural Networks (GNNs) have become a dominant paradigm for representation learning on graph-structured data. Early spectral and spatial methods, such as GCN~\citep{kipf2016semi}, GraphSAGE~\citep{hamilton2017inductive}, GAT~\citep{velivckovic2017graph}, and SGC~\citep{wu2019simplifying}, learn node representations by recursively aggregating information from local neighborhoods. These models have achieved strong performance on node classification, link prediction, and graph classification tasks by exploiting graph topology as an inductive bias. Subsequent studies further extend GNNs to heterogeneous/heterophilic graphs, graph transformers, and scalable graph learning settings~\citep{hu2020heterogeneous,wang2019heterogeneous,rampavsek2022recipe,zhu2020beyond,xue2024haste,xue2025h,xue2025visagnn}. Despite their success, most conventional GNNs assume that the observed topology is reliable and that node attributes are uni-modal. This assumption is often violated in practical graphs, where edges may be noisy or incomplete with respect to multimodal semantics, and node attributes come from various modalities such as text and vision. Directly applying standard GNNs to concatenated multimodal features or fixed graph structures may therefore propagate modality-specific noise and fail to capture cross-modal dependencies during message passing.

\textbf{Multimodal Representation Learning.}
Multimodal representation learning has been a central topic in recent years.
It aims to encode signals from different modalities, such as text, images, audio, and video, into semantically aligned embedding spaces. Contrastive vision-language models, such as CLIP~\citep{radford2021learning} and ALIGN~\citep{jia2021scaling}, learn transferable representations by aligning paired images and texts through large-scale contrastive pre-training. Later models further improve multimodal understanding and generation through stronger vision encoders, language models, cross-attention modules, and instruction tuning, including BLIP~\citep{li2022blip,li2023blip}, ImageBind~\citep{girdhar2023imagebind}, LLaVA~\citep{liu2023visual}, and Qwen-VL~\citep{bai2023qwen}. These models provide powerful feature extractors and have been widely used to initialize multimodal node features. However, most multimodal representation learning methods focus on instance-level alignment and do not explicitly model graph topology. As a result, their pretrained embeddings are usually static, task-agnostic, and insensitive to downstream graph structures. In multimodal graphs, semantic relationships between nodes are jointly determined by modality content and graph connectivity. Therefore, simply using pretrained multimodal embeddings is insufficient: the model should further adapt modality alignment and information propagation according to graph topology and task supervision.

\textbf{Multimodal Graph Learning.}
Most previous works primarily focus on a single modality, such as text-attributed graphs~\citep{chien2021node,zhao2022learning,xue2023efficient}.
Recntly, multimodal graph learning extends graph representation learning to graphs whose nodes or edges are associated with multiple modalities. A straightforward solution is to concatenate multimodal features and feed them into conventional GNNs, or to encode each modality with separate GNN branches followed by late fusion. Representative multimodal graph methods, such as MMGCN and MGAT~\citep{zhu2025mosaic}, exploit modality-specific representations and attention-based fusion to combine heterogeneous signals. More recent studies attempt to build more general multimodal graph learning frameworks. UniGraph2~\citep{he2025unigraph2} builds a graph foundation modal to learn a unified embedding space for multimodal graphs using modality-specific encoders, GNNs, and mixture-of-experts alignment across domains and modalities. NSG-MoE~\citep{zhang2026modality} treats modality as heterogeneity by splitting each node into modality-specific sub-nodes and routing heterogeneous message flows through structured experts. LION~\citep{li2026lion} introduces a Clifford-algebra-based paradigm to perform modality alignment and adaptive holographic aggregation on multimodal-attributed graphs. PLANET~\citep{liu2026toward} further studies multimodal graph foundation models and proposes topology-aware modality interaction and alignment through embedding-wise domain gating and discretized semantic representation spaces. These methods demonstrate the importance of multimodal interaction and alignment in graph learning. However, many existing approaches either rely on fixed observed topology, perform modality interaction after independent propagation, or emphasize representation alignment without explicitly modeling the reciprocal relationship between graph structure and multimodal semantics, thereby limiting the effectiveness of graph-modality fusion and cross-modal alignment.

\textbf{Positioning of ReCoG.}
Different from prior work, 
our approach treats multimodal graph learning as a coupled topology-semantics co-evolution problem. First, we introduce a multimodal graph refiner that adaptively reweights original edges and adds semantic neighbors using joint textual and visual evidence, addressing the topology--semantics mismatch in the original graph. Second, we perform iterative cross-modal message passing over the refined topology, allowing textual and visual representations to align and exchange information during propagation rather than only at the final fusion stage. This design jointly improves graph structure and multimodal representations in an end-to-end manner, providing a more task-adaptive alternative to fixed-topology GNNs, static multimodal encoders, and late-fusion multimodal graph models.


\section{Results for Link Prediction}
\label{app:linkpred}
We provide the full link prediction results in Table~\ref{tab:lp-main}. A comprehensive analysis is presented in Section~\ref{sec:lp-results}.

\begin{table*}[t]
\caption{Link prediction performance on Amazon-Sports and Amazon-Cloth.~\citep{zhu2025mosaic}. The \textcolor{orange}{Best} and \textcolor{teal}{runner-up} are highlighted.}
\centering
\footnotesize
\setlength{\tabcolsep}{5pt}
\renewcommand{\arraystretch}{1.15}
\begin{tabular}{l l c c c c c c}
\toprule
 & \textbf{Encoder}
 & \multicolumn{3}{c}{\textbf{Amazon-Sports}}
 & \multicolumn{3}{c}{\textbf{Amazon-Cloth}} \\
\cmidrule(lr){3-5}\cmidrule(lr){6-8}
 & & \textbf{MRR} & \textbf{H@1} & \textbf{H@10}
   & \textbf{MRR} & \textbf{H@1} & \textbf{H@10} \\
\midrule

\multirow{4}{*}{\textbf{MMGCN}}
 & CLIP      & 31.96\,$\pm$\,0.10 & 16.35\,$\pm$\,0.11 & 68.46\,$\pm$\,0.08 & 22.20\,$\pm$\,0.05 & 10.76\,$\pm$\,0.10 & 46.62\,$\pm$\,0.12 \\
 & ViT+T5    & 30.33\,$\pm$\,0.08 & 15.01\,$\pm$\,0.05 & 66.41\,$\pm$\,0.11 & 19.45\,$\pm$\,0.34 & 9.22\,$\pm$\,0.20  & 40.49\,$\pm$\,0.61 \\
 & ImageBind & 31.74\,$\pm$\,0.21 & 16.45\,$\pm$\,0.13 & 67.39\,$\pm$\,0.74 & 24.72\,$\pm$\,0.19 & 12.47\,$\pm$\,0.09 & 51.32\,$\pm$\,0.56 \\
 & DINO+T5   & 30.04\,$\pm$\,0.27 & 14.98\,$\pm$\,0.07 & 64.56\,$\pm$\,0.56 & 21.77\,$\pm$\,0.23 & 10.47\,$\pm$\,0.12 & 45.81\,$\pm$\,0.52 \\
\midrule

\multirow{4}{*}{\textbf{MGAT}}
 & CLIP      & 27.56\,$\pm$\,0.30 & 13.55\,$\pm$\,0.29 & 60.21\,$\pm$\,0.21 & 21.38\,$\pm$\,0.23 & 10.39\,$\pm$\,0.22 & 44.60\,$\pm$\,0.36 \\
 & ViT+T5    & 30.15\,$\pm$\,0.34 & 15.28\,$\pm$\,0.34 & 64.84\,$\pm$\,0.41 & 20.59\,$\pm$\,0.41 & 9.79\,$\pm$\,0.30  & 43.44\,$\pm$\,0.76 \\
 & ImageBind & 30.15\,$\pm$\,0.12 & 15.50\,$\pm$\,0.05 & 64.20\,$\pm$\,0.43 & 22.13\,$\pm$\,0.27 & 10.96\,$\pm$\,0.15 & 45.84\,$\pm$\,0.57 \\
 & DINO+T5   & 28.91\,$\pm$\,0.09 & 14.47\,$\pm$\,0.18 & 62.11\,$\pm$\,0.22 & 21.42\,$\pm$\,0.13 & 10.38\,$\pm$\,0.13 & 44.11\,$\pm$\,0.50 \\
\midrule

\multirow{4}{*}{\textbf{GCN}}
 & CLIP      & 31.38\,$\pm$\,0.08 & 16.58\,$\pm$\,0.13 & 66.14\,$\pm$\,0.08 & 22.28\,$\pm$\,0.05 & 11.83\,$\pm$\,0.04 & 43.52\,$\pm$\,0.10 \\
 & ViT+T5    & 30.83\,$\pm$\,0.07 & 16.31\,$\pm$\,0.08 & 64.76\,$\pm$\,0.15 & 21.60\,$\pm$\,0.05 & 11.37\,$\pm$\,0.05 & 42.29\,$\pm$\,0.14 \\
 & ImageBind & 31.67\,$\pm$\,0.09 & 17.07\,$\pm$\,0.14 & 65.61\,$\pm$\,0.10 & 22.81\,$\pm$\,0.02 & 12.27\,$\pm$\,0.05 & 44.28\,$\pm$\,0.09 \\
 & DINO+T5   & 30.42\,$\pm$\,0.02 & 16.02\,$\pm$\,0.03 & 64.02\,$\pm$\,0.06 & 21.19\,$\pm$\,0.04 & 11.09\,$\pm$\,0.06 & 41.46\,$\pm$\,0.16 \\
\midrule

\multirow{4}{*}{\textbf{SAGE}}
 & CLIP      & 33.83\,$\pm$\,0.08 & 17.57\,$\pm$\,0.14 & 71.90\,$\pm$\,0.07 & 24.58\,$\pm$\,0.13 & 12.16\,$\pm$\,0.11 & 51.12\,$\pm$\,0.09 \\
 & ViT+T5    & 32.01\,$\pm$\,0.10 & 15.94\,$\pm$\,0.17 & 69.84\,$\pm$\,0.21 & 23.11\,$\pm$\,0.05 & 11.10\,$\pm$\,0.04 & 48.89\,$\pm$\,0.09 \\
 & ImageBind & 34.32\,$\pm$\,0.11 & 17.87\,$\pm$\,0.23 & 73.04\,$\pm$\,0.15 & 25.20\,$\pm$\,0.09 & 12.63\,$\pm$\,0.05 & 52.53\,$\pm$\,0.21 \\
 & DINO+T5   & 32.20\,$\pm$\,0.12 & 16.19\,$\pm$\,0.20 & 69.98\,$\pm$\,0.32 & 22.98\,$\pm$\,0.01 & 11.12\,$\pm$\,0.04 & 48.28\,$\pm$\,0.11 \\
\midrule

\multirow{4}{*}{\textbf{MLP}}
 & CLIP      & 28.22\,$\pm$\,0.09 & 14.54\,$\pm$\,0.16 & 59.40\,$\pm$\,0.08 & 21.10\,$\pm$\,0.04 & 10.70\,$\pm$\,0.05 & 42.77\,$\pm$\,0.05 \\
 & ViT+T5    & 24.81\,$\pm$\,0.05 & 11.63\,$\pm$\,0.05 & 54.78\,$\pm$\,0.04 & 17.65\,$\pm$\,0.06 & 8.14\,$\pm$\,0.04  & 36.77\,$\pm$\,0.06 \\
 & ImageBind & 30.45\,$\pm$\,0.14 & 15.91\,$\pm$\,0.10 & 64.10\,$\pm$\,0.07 & 22.18\,$\pm$\,0.02 & 11.42\,$\pm$\,0.04 & 44.86\,$\pm$\,0.06 \\
 & DINO+T5   & 24.81\,$\pm$\,0.16 & 11.62\,$\pm$\,0.18 & 54.97\,$\pm$\,0.22 & 17.53\,$\pm$\,0.11 & 8.07\,$\pm$\,0.09  & 36.53\,$\pm$\,0.26 \\
\midrule

\multirow{4}{*}{\shortstack[l]{\textbf{ReCoG-sim} \\(GCN)}}
 & CLIP      & \cellcolor{teal!30}38.53\,$\pm$\,0.03 & 21.90\,$\pm$\,0.05 & 75.11\,$\pm$\,0.02 & 28.30\,$\pm$\,0.03 & 15.64\,$\pm$\,0.06 & 54.96\,$\pm$\,0.05 \\
 & ViT+T5    & 36.82\,$\pm$\,0.05 & 20.49\,$\pm$\,0.03 & 73.36\,$\pm$\,0.05 & 24.66\,$\pm$\,0.05 & 12.94\,$\pm$\,0.09 & 49.19\,$\pm$\,0.05 \\
 & ImageBind & 38.36\,$\pm$\,0.02 & \cellcolor{teal!30}21.95\,$\pm$\,0.03 & 74.68\,$\pm$\,0.09 & \cellcolor{teal!30}28.60\,$\pm$\,0.06 & \cellcolor{orange!50}15.89\,$\pm$\,0.08 & \cellcolor{teal!30}55.71\,$\pm$\,0.08 \\
 & DINO+T5   & 37.16\,$\pm$\,0.05 & 21.29\,$\pm$\,0.03 & 73.31\,$\pm$\,0.02 & 27.10\,$\pm$\,0.06 & 14.51\,$\pm$\,0.06 & 54.02\,$\pm$\,0.16 \\
\midrule

\multirow{4}{*}{\shortstack[l]{\textbf{ReCoG-sim} \\(SAGE)}}
 & CLIP      & 35.34\,$\pm$\,0.03 & 18.66\,$\pm$\,0.12 & 74.23\,$\pm$\,0.08 & 27.02\,$\pm$\,0.09 & 14.14\,$\pm$\,0.18 & 54.38\,$\pm$\,0.08 \\
 & ViT+T5    & 34.55\,$\pm$\,0.18 & 18.31\,$\pm$\,0.18 & 71.11\,$\pm$\,0.10 & 25.38\,$\pm$\,0.11 & 12.99\,$\pm$\,0.09 & 51.78\,$\pm$\,0.08 \\
 & ImageBind & 36.90\,$\pm$\,0.08 & 20.41\,$\pm$\,0.06 & 75.03\,$\pm$\,0.08 & 27.08\,$\pm$\,0.06 & 14.48\,$\pm$\,0.11 & 54.03\,$\pm$\,0.12 \\
 & DINO+T5   & 35.36\,$\pm$\,0.10 & 19.13\,$\pm$\,0.11 & 72.40\,$\pm$\,0.03 & 26.36\,$\pm$\,0.03 & 14.00\,$\pm$\,0.05 & 53.05\,$\pm$\,0.03 \\
\midrule

\multirow{4}{*}{\shortstack[l]{\textbf{ReCoG-res} \\(GCN)}}
 & CLIP      & \cellcolor{orange!50}38.57\,$\pm$\,0.03 & \cellcolor{orange!50}22.04\,$\pm$\,0.05 & \cellcolor{teal!30}75.37\,$\pm$\,0.02 & 28.35\,$\pm$\,0.06 & 15.64\,$\pm$\,0.08 & 54.60\,$\pm$\,0.05 \\
 & ViT+T5    & 37.28\,$\pm$\,0.03 & 20.89\,$\pm$\,0.03 & 74.06\,$\pm$\,0.06 & 27.86\,$\pm$\,0.05 & 15.32\,$\pm$\,0.09 & 54.35\,$\pm$\,0.05 \\
 & ImageBind & 38.42\,$\pm$\,0.09 & 21.73\,$\pm$\,0.05 & \cellcolor{orange!50}75.60\,$\pm$\,0.03 & \cellcolor{orange!50}28.73\,$\pm$\,0.05 & \cellcolor{teal!30}15.75\,$\pm$\,0.08 & \cellcolor{orange!50}55.85\,$\pm$\,0.08 \\
 & DINO+T5   & 37.39\,$\pm$\,0.06 & 21.29\,$\pm$\,0.03 & 74.89\,$\pm$\,0.01 & 27.78\,$\pm$\,0.06 & 15.11\,$\pm$\,0.05 & 54.54\,$\pm$\,0.03 \\
\midrule

\multirow{4}{*}{\shortstack[l]{\textbf{ReCoG-res} \\(SAGE)}}
 & CLIP      & 36.22\,$\pm$\,0.03 & 19.92\,$\pm$\,0.12 & 73.74\,$\pm$\,0.08 & 27.37\,$\pm$\,0.09 & 14.49\,$\pm$\,0.12 & 55.37\,$\pm$\,0.02 \\
 & ViT+T5    & 35.07\,$\pm$\,0.10 & 18.64\,$\pm$\,0.18 & 73.40\,$\pm$\,0.10 & 26.55\,$\pm$\,0.10 & 13.72\,$\pm$\,0.09 & 54.19\,$\pm$\,0.05 \\
 & ImageBind & 36.13\,$\pm$\,0.08 & 19.86\,$\pm$\,0.06 & 73.97\,$\pm$\,0.12 & 27.62\,$\pm$\,0.06 & 14.76\,$\pm$\,0.11 & 55.37\,$\pm$\,0.06 \\
 & DINO+T5   & 35.60\,$\pm$\,0.05 & 18.81\,$\pm$\,0.10 & 73.19\,$\pm$\,0.05 & 25.95\,$\pm$\,0.08 & 13.96\,$\pm$\,0.05 & 54.08\,$\pm$\,0.03 \\
\bottomrule
\end{tabular}
\label{tab:lp-main}
\end{table*}

\section{Top-K Selection (Eqns.~\eqref{eq:sim_topk} and~\eqref{eqn:res_topk}).}
\label{app:topk}

\begin{table}[h]
\centering
\caption{Ablation study on the number of candidate neighbors $K$ in the graph refiner. 
}
\label{tab:topk}
\resizebox{0.9\linewidth}{!}{
\begin{tabular}{cl cc cc}
\toprule
\multirow{2}{*}{Refiner} & \multirow{2}{*}{$K$} & \multicolumn{2}{c}{Node Classification (\%)} & \multicolumn{2}{c}{Link Prediction (MRR \%)} \\
\cmidrule(lr){3-4} \cmidrule(lr){5-6}
 & & Ele-Fashion & Goodreads & Amazon-Sports & Amazon-Clothing \\
\midrule
\multirow{5}{*}{\rotatebox[origin=c]{90}{sim}}
& 1  & 87.60 \,$\pm$\,0.05 & 89.65 \,$\pm$\,0.05 & 36.72 \,$\pm$\,0.11 & 27.07 \,$\pm$\,0.10\\
 & 3  & 88.17 \,$\pm$\,0.03 & 90.43 \,$\pm$\,0.09 & 37.81 \,$\pm$\,0.06 & 27.93 \,$\pm$\,0.07\\
 & 5  & 88.56 \,$\pm$\,0.05 & 90.31 \,$\pm$\,0.05 & 38.53 \,$\pm$\,0.03 & 28.30 \,$\pm$\,0.03 \\
 & 8  & 88.53 \,$\pm$\,0.05 & 90.16 \,$\pm$\,0.06 & 38.32 \,$\pm$\,0.09 & 28.01 \,$\pm$\,0.06\\
 & 12  & 87.95 \,$\pm$\,0.06 & 89.43 \,$\pm$\,0.10 & 37.23 \,$\pm$\,0.10 & 26.68 \,$\pm$\,0.05\\
\midrule
\multirow{5}{*}{\rotatebox[origin=c]{90}{res}}
 & 1  & 87.88 \,$\pm$\,0.08 & 89.65 \,$\pm$\,0.08 & 37.36 \,$\pm$\,0.09 & 27.37 \,$\pm$\,0.07 \\
 & 3  & 88.30 \,$\pm$\,0.06 & 90.68 \,$\pm$\,0.06 & 38.51 \,$\pm$\,0.05 & 28.35 \,$\pm$\,0.03\\
 & 5  & 88.60 \,$\pm$\,0.08 & 90.59 \,$\pm$\,0.05 & 38.57 \,$\pm$\,0.03 & 28.25 \,$\pm$\,0.02\\
 & 8  & 88.38 \,$\pm$\,0.03 & 90.66 \,$\pm$\,0.03 & 38.51 \,$\pm$\,0.08 & 28.02 \,$\pm$\,0.05\\
& 12  & 87.85 \,$\pm$\,0.08 & 90.07 \,$\pm$\,0.05 & 37.70 \,$\pm$\,0.08 & 26.91 \,$\pm$\,0.06\\
\bottomrule
\end{tabular}
}
\end{table}

Table~\ref{tab:topk} examines the sensitivity of both the Residual and Cosine refiners to the number of candidate neighbors $K$. We use CLIP as the encoder and GCN as the backbone for all experiments. Across all datasets and both refiners, we observe a consistent pattern: performance improves as $K$ increases from 1 to around 3--5, and then gradually degrades when $K=12$.

When $K=1$, the refiner has too few candidates to meaningfully adjust the graph topology, leading to weaker performance. Conversely, an overly large $K$ may introduce semantically distant nodes as candidates, injecting noise into the refined adjacency and diluting the quality of the added edges. Notably, performance remains relatively stable in the range of $K=3$ to $K=5$, where the candidate pool is large enough to capture missing semantic neighbors while remaining small enough to avoid spurious connections. Based on these results, we set $K=3$ or $K=5$ as the default in all experiments for simplicity.

\section{Complexity Optimization of Cross Modal Message Passing}
\label{app:opt}

We describe two successive optimizations that reduce computational cost without altering the output.

\subsection{Optimization I: Supergraph Reformulation}

Directly constructing $\mathbf{A}_{\text{sup}}$ requires $2N$ nodes and $4E + 2N$ edges. By computing each term in Eq.~\eqref{eq:cross_modal} independently, we avoid constructing the supergraph entirely. This decomposition requires four \texttt{conv} calls (two per modality), each operating on $N$ nodes and $E$ edges, plus two \texttt{lin} calls at $O(N \cdot d_h^2)$ each. The total per-layer cost is:
\begin{equation}
    4 \times \underbrace{O(E \cdot d_h + N \cdot d_h^2)}_{\texttt{conv}} + 2 \times \underbrace{O(N \cdot d_h^2)}_{\texttt{lin}}
\end{equation}
compared to the supergraph cost of $O((4E + 2N) \cdot d_h + 2N \cdot d_h^2)$ for a single \texttt{conv} over $2N$ nodes. The decomposed version halves the node count per call and eliminates the overhead of constructing the supergraph edge index (node ID offsets, feature concatenation, and additional $2N$ cross-modal identity edges).

\subsection{Optimization II: Linear Merging of Terms 1 and 3}

For GNN layers whose aggregation operator is linear in the input features, Terms 1 and 3 can be merged into a single \texttt{conv} call:
\begin{equation}
    \texttt{conv}(\mathbf{h}_t^{(l)},\, \mathbf{A}) + \rho \cdot \texttt{conv}(\mathbf{h}_v^{(l)},\, \mathbf{A}) 
    = \texttt{conv}\!\big(\mathbf{h}_t^{(l)} + \rho \cdot \mathbf{h}_v^{(l)},\, \mathbf{A}\big)
    \label{eq:merged}
\end{equation}

Without merging, computing Terms 1 and 3 for one modality requires two \texttt{conv} calls:
\begin{equation}
    \underbrace{\texttt{conv}(\mathbf{h}_t^{(l)},\, \mathbf{A})}_{\text{Term 1}} + \rho \cdot \underbrace{\texttt{conv}(\mathbf{h}_v^{(l)},\, \mathbf{A})}_{\text{Term 3}}
    \quad\Rightarrow\quad 2 \times O(E \cdot d_h) + 2 \times O(N \cdot d_h^2)
\end{equation}
After merging, the cost reduces to a single element-wise addition followed by one \texttt{conv} call:
\begin{equation}
    \underbrace{\mathbf{h}_t^{(l)} + \rho \cdot \mathbf{h}_v^{(l)}}_{O(N \cdot d_h)} \;\longrightarrow\; \underbrace{\texttt{conv}(\,\cdot\,,\, \mathbf{A})}_{O(E \cdot d_h + N \cdot d_h^2)}
\end{equation}
The $O(N \cdot d_h)$ addition is negligible compared to the \texttt{conv} cost, so merging effectively halves both the aggregation and linear transformation costs per modality. The complete per-layer update becomes:
\begin{align}
    \mathbf{h}_t^{(l+1)} &= \text{LN}\!\Big(\texttt{conv}\!\big(\mathbf{h}_t^{(l)} + \rho\,\mathbf{h}_v^{(l)},\; \mathbf{A}\big) + \gamma \cdot \texttt{lin}(\mathbf{h}_v^{(l)})\Big) \label{eq:final-t} \\
    \mathbf{h}_v^{(l+1)} &= \text{LN}\!\Big(\texttt{conv}\!\big(\mathbf{h}_v^{(l)} + \rho\,\mathbf{h}_t^{(l)},\; \mathbf{A}\big) + \gamma \cdot \texttt{lin}(\mathbf{h}_t^{(l)})\Big) \label{eq:final-v}
\end{align}
requiring two \texttt{conv} calls and two \texttt{lin} calls per layer, yielding the total per-layer cost of $O(E \cdot d_h + N \cdot d_h^2)$ reported in Table~\ref{tab:complexity}.

\paragraph{Applicability.} The merging in Eq.~\eqref{eq:merged} requires linearity of the aggregation operator and holds for GCN, GraphSAGE-mean, and other linear-aggregation GNNs. For attention-based architectures such as GAT, where the aggregation weights depend nonlinearly on the input features, the merging does not hold exactly. In this case, we fall back to the decomposed implementation (four \texttt{conv} calls), which still avoids constructing the supergraph.

\section{Topology–semantics Misalignment}
\label{app:misalign}

Let the observed multimodal graph be denoted as $\mathcal{G} = (\mathcal{V}, \mathcal{E})$,
where $\mathcal{V}$ is the node set and $\mathcal{E}$ is the observed edge set. Each node $i \in \mathcal{V}$ is associated with multimodal features
$\mathbf{x}_i = [\mathbf{x}_i^{t} \, \| \, \mathbf{x}_i^{v}]$,
where $\mathbf{x}_i^{t} \in \mathbb{R}^{d_t}$ and $\mathbf{x}_i^{v} \in \mathbb{R}^{d_v}$ denote textual and visual features, respectively. Denote by $\mathbf{A}_{\text{orig}}$ the original adjacency matrix, $\mathbf{A}_{\text{orig}}(i,j)=1$ if $(i,j)\in \mathcal{E}$, $0$ otherwise.

A fundamental challenge in multimodal graphs is that the observed topology is only partially aligned with multimodal semantics. 
To examine this issue, we analyze the multimodal semantic similarity distributions on Reddit-M (see Sec.~\ref{sec:prelim}) and Movies~\citep{yan2025graph}. 
For a node pair $(i,j)$, we compute the textual and visual cosine similarities as
\begin{equation}
    S_T(i,j) = \mathrm{cos}(\mathbf{x}^t_i, \mathbf{x}^t_j),
    \qquad
    S_V(i,j) = \mathrm{cos}(\mathbf{x}^v_i, \mathbf{x}^v_j),
    \label{eqn:cosine}
\end{equation}
For simplicity, we plot the multimodal semantic similarity as the average
\begin{equation}
S_{MM}(i,j) = \frac{1}{2}\big(S_T(i,j) + S_V(i,j)\big),
\end{equation}
where the encoded features are obtained from CLIP.

\begin{figure}[h]
\centering
\includegraphics[width=\linewidth]{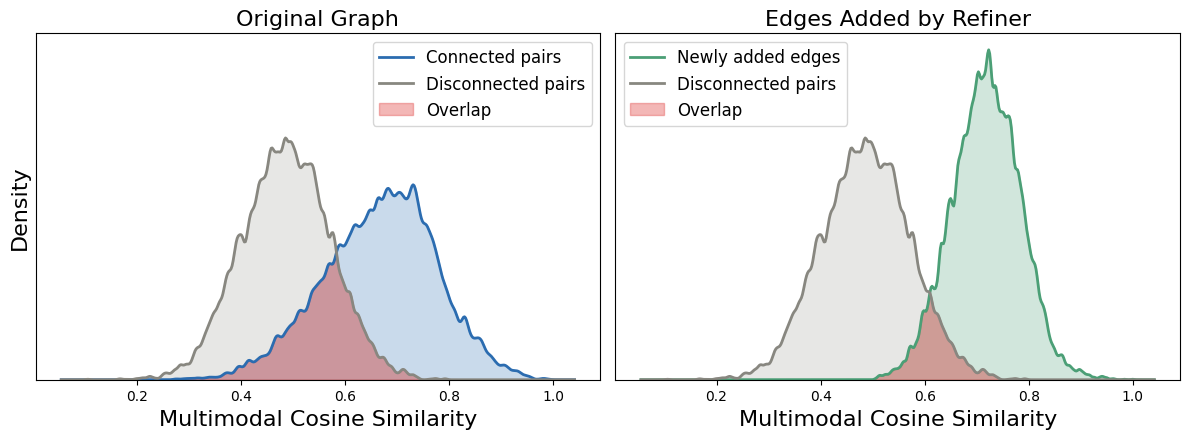}
\caption{Topology--semantics alignment.}
\label{fig:refinement2}
\end{figure}

As shown on the left side of both Fig.~\ref{fig:refinement} and Fig.~\ref{fig:refinement2}, we compare the $S_{MM}$ distributions of connected and disconnected node pairs in the original graph. Connected pairs generally exhibit higher multimodal similarity, indicating that the original topology provides useful structural priors. However, the two distributions show considerable overlap, revealing two types of topology--semantics discrepancy: 
\textit{(i)} some connected pairs have low multimodal similarity, suggesting the presence of noisy edges; 
\textit{(ii)} some disconnected pairs have high multimodal similarity, indicating missing semantic neighbors.

The right side of both Fig.~\ref{fig:refinement} (Cosine Similarity Refiner) and Fig.~\ref{fig:refinement2} (Residual Refiner) illustrates the effect of the refiner. The newly added edges concentrate in a higher similarity range compared to the original connected pairs, and their overlap with the disconnected distribution is substantially reduced. This confirms that the refiner selectively recovers semantically meaningful missing edges without introducing topological noise.

These observations motivate a topology refinement mechanism that preserves the observed connectivity as a structural prior while adaptively suppressing noisy edges and recovering missing semantic neighbors based on joint textual and visual evidence.

\section{Complexity Analysis.}
\label{app:complexity}
We analyze the per-batch computational complexity of ReCoG (Table~\ref{tab:complexity}). Let $N$ denote the number of nodes in the sampled subgraph, $B$ the number of seed nodes, $E$ the number of edges, $k$ the number of new edges per seed, $r$ the candidate ratio, $d = d_t + d_v$ the raw feature dimension, $d_h$ the hidden dimension in the projection at coarse  filtering, and $L$ the number of GNN layers.

\begin{table}[h]
\centering
\caption{Per-batch computational complexity of ReCoG major components.}
\label{tab:complexity}
\begin{tabular}{@{}ll@{}}
\toprule
\textbf{Component} & \textbf{Complexity} \\
\midrule
\quad Cosine similarity ($S_T$, $S_V$) & $O(BNd)$ \\
\midrule
\quad Original edge scoring (MLP) & $O(E \cdot d \cdot d_h)$ \\
\quad Coarse filtering & $O(BNd_h)$ \\
\quad Fine scoring (MLP) & $O(Brk \cdot d \cdot d_h)$ \\
\midrule
\quad Cross-modal message passing ($\times L$) & $O(L(E \cdot d_h + N \cdot d_h^2))$ \\
\bottomrule
\end{tabular}
\end{table}

\section{Proof of Proposition~\ref{prop:refiner} (Strict Generalization of Residual Refinement)}
\label{app:proof_1}
 
We prove $\mathcal{F}_{\mathrm{sim}} \subsetneq \mathcal{F}_{\mathrm{res}}$ by establishing inclusion and then strictness. Recall that the similarity refiner induces the edge-scoring family
\begin{equation}
    g_{\mathrm{sim}}(i,j)
    = \alpha\, A_{\mathrm{orig}}(i,j)
    + (1-\alpha)\bigl[\lambda\, S_V(i,j) + (1-\lambda)\, S_T(i,j)\bigr],
    \quad (i,j)\in\mathcal{C},
    \label{eq:g_sim}
\end{equation}
with global coefficients $\alpha,\lambda\in[0,1]$, while the residual refiner induces
\begin{equation}
    g_{\mathrm{res}}(i,j)
    = A_{\mathrm{orig}}(i,j)
    + \delta\cdot\tanh\!\bigl(f_\theta(\phi(i,j))\bigr),
    \quad (i,j)\in\mathcal{C},
    \label{eq:g_res}
\end{equation}
where $\phi(i,j)$ is the multimodal pair feature (Eq.~\eqref{eq:phi}) and $f_\theta$ is an MLP with sufficient width.
 
\paragraph{Part 1: $\mathcal{F}_{\mathrm{sim}} \subseteq \mathcal{F}_{\mathrm{res}}$.}
Take any $g_{\mathrm{sim}} \in \mathcal{F}_{\mathrm{sim}}$.  For each $(i,j)\in\mathcal{C}$, define the target value
\begin{equation}
    u(i,j) = \operatorname{arctanh}\!\Bigl(\tfrac{g_{\mathrm{sim}}(i,j) - A_{\mathrm{orig}}(i,j)}{\delta}\Bigr).
    \label{eq:u_def}
\end{equation}
By the range assumption and $|g_{\mathrm{sim}}(i,j) - A_{\mathrm{orig}}(i,j)| < \delta$, the argument of $\operatorname{arctanh}$ lies in $(-1,1)$, so $u(i,j)$ is well-defined and finite for every $(i,j) \in \mathcal{C}$.
 
The similarity-based score depends continuously on $(S_T(i,j), S_V(i,j))$, which are themselves continuous functions of the pair features $\phi(i,j)$ on the fixed candidate set $\mathcal{C}$.  Since $\operatorname{arctanh}$ is continuous on $(-1,1)$, the composite map $\phi(i,j) \mapsto u(i,j)$ is continuous on $\mathcal{C}$.
 
By the universal approximation theorem for MLPs with sufficient width, there exists an MLP $f_\theta$ such that
\begin{equation}
    \bigl|f_\theta(\phi(i,j)) - u(i,j)\bigr| < \epsilon, \quad \forall (i,j) \in \mathcal{C},
\end{equation}
for any $\epsilon > 0$.  Substituting into the residual scorer:
\begin{align}
    g_{\mathrm{res}}(i,j)
    &= A_{\mathrm{orig}}(i,j) + \delta \cdot \tanh\!\bigl(f_\theta(\phi(i,j))\bigr) \\
    &\approx A_{\mathrm{orig}}(i,j) + \delta \cdot \tanh\!\bigl(u(i,j)\bigr) \\
    &= A_{\mathrm{orig}}(i,j) + \delta \cdot \frac{g_{\mathrm{sim}}(i,j) - A_{\mathrm{orig}}(i,j)}{\delta} \\
    &= g_{\mathrm{sim}}(i,j).
\end{align}
Since the approximation error can be made arbitrarily small, every function in $\mathcal{F}_{\mathrm{sim}}$ can be realized (or approximated to arbitrary precision) by a function in $\mathcal{F}_{\mathrm{res}}$, establishing $\mathcal{F}_{\mathrm{sim}} \subseteq \mathcal{F}_{\mathrm{res}}$.
 
\paragraph{Part 2: Strictness ($\mathcal{F}_{\mathrm{sim}} \neq \mathcal{F}_{\mathrm{res}}$).}
Every function in $\mathcal{F}_{\mathrm{sim}}$ depends on a node pair $(i,j)$ only through the two scalars $(S_T(i,j), S_V(i,j))$, combined via a global linear rule with shared coefficients $\alpha$ and $\lambda$. Therefore, for any two pairs $(i,j)$ and $(u,v)$ satisfying
\[
S_T(i,j)=S_T(u,v), \quad S_V(i,j)=S_V(u,v),
\quad A_{\mathrm{orig}}(i,j)=A_{\mathrm{orig}}(u,v),
\]
every $g_{\mathrm{sim}}\in\mathcal{F}_{\mathrm{sim}}$ must assign them the same score:
\begin{equation}
    g_{\mathrm{sim}}(i,j) = g_{\mathrm{sim}}(u,v).
    \label{eq:sim_equal}
\end{equation}
 
Now consider the residual scorer.  Its input is the full pair feature $\phi(i,j)$ (Eq.~\eqref{eq:phi}), which includes concatenated raw features, element-wise Hadamard products $\mathbf{x}_i^t \odot \mathbf{x}_j^t$, $\mathbf{x}_i^v \odot \mathbf{x}_j^v$, and absolute differences $|\mathbf{x}_i^t - \mathbf{x}_j^t|$, $|\mathbf{x}_i^v - \mathbf{x}_j^v|$.
 
We construct a concrete counterexample.  Let $d_t = 2$ and consider two node pairs:
\begin{align}
    \text{Pair 1:} &\quad \mathbf{x}_i^t = (1, 0), \quad \mathbf{x}_j^t = (0, 1), \\
    \text{Pair 2:} &\quad \mathbf{x}_u^t = \tfrac{1}{\sqrt{2}}(1, 1), \quad \mathbf{x}_v^t = \tfrac{1}{\sqrt{2}}(1, -1).
\end{align}
Both pairs have unit-norm vectors with $S_T = \mathbf{x}_i^t \cdot \mathbf{x}_j^t = 0$ and $S_T = \mathbf{x}_u^t \cdot \mathbf{x}_v^t = 0$. Assigning identical vision features to both pairs ensures $S_V$ is also equal. 
 
However, the Hadamard products differ:
\begin{align}
    \mathbf{x}_i^t \odot \mathbf{x}_j^t &= (0, 0), \\
    \mathbf{x}_u^t \odot \mathbf{x}_v^t &= \tfrac{1}{2}(1, -1),
\end{align}
so $\phi(i,j) \neq \phi(u,v)$.  Since an MLP can realize non-linear, dimension-weighted functions of $\phi$, there exists $g_{\mathrm{res}} \in \mathcal{F}_{\mathrm{res}}$ such that
\begin{equation}
    g_{\mathrm{res}}(i,j) \neq g_{\mathrm{res}}(u,v),
\end{equation}
which is impossible for any $g_{\mathrm{sim}} \in \mathcal{F}_{\mathrm{sim}}$ by Eq.~\eqref{eq:sim_equal}.
 
Therefore $\mathcal{F}_{\mathrm{sim}} \subsetneq \mathcal{F}_{\mathrm{res}}$. \qed

\begin{remark}
\label{app:remark1}
Proposition~\ref{prop:refiner} characterizes \emph{expressiveness}, 
not generalization. The inclusion follows from a universal-approximation 
argument on the residual MLP, and strictness from dimension-wise 
interactions (e.g., $\mathbf{x}_i^t \odot \mathbf{x}_j^t$) that 
cosine similarity collapses to a scalar. Whether this additional capacity 
translates to better task performance depends on the learning context: 
our ablations (Sec.~\ref{sec:ablation}) show that the residual 
refiner's advantage emerges only when coupled with cross-modal propagation.
\end{remark}

\section{Proof of Proposition~\ref{prop:joint} (Joint Optimization Contains Non-Reciprocal Alternatives)}
\label{app:proof_3}

Let $\Theta_{\mathrm{ref}}$ and $\Theta_{\mathrm{gnn}}$ denote the parameter spaces of the refiner and the GNN, respectively. Any fixed refiner $\theta_{\mathrm{ref}}^{\mathrm{fix}}$ restricts the optimization to the slice
$\{\theta_{\mathrm{ref}}^{\mathrm{fix}}\}\times\Theta_{\mathrm{gnn}}$, which is a subset of the full feasible set
$\Theta_{\mathrm{ref}}\times\Theta_{\mathrm{gnn}}$. Therefore, the infimum over the full joint space cannot exceed the infimum over this restricted slice:
\begin{equation}
    \inf_{(\theta_{\mathrm{ref}},\,\theta_{\mathrm{gnn}})
    \in
    \Theta_{\mathrm{ref}}\times\Theta_{\mathrm{gnn}}}
    \mathcal{J}(\theta_{\mathrm{ref}},\theta_{\mathrm{gnn}})
    \le
    \inf_{\theta_{\mathrm{gnn}}\in\Theta_{\mathrm{gnn}}}
    \mathcal{J}(\theta_{\mathrm{ref}}^{\mathrm{fix}},\theta_{\mathrm{gnn}}).
\end{equation}
This includes the identity refiner $A_{\mathrm{refine}}=A_{\mathrm{orig}}$ when it belongs to the refiner parameter space, as well as any frozen two-stage refiner. \qed
 
 
\section{Proof of Theorem~\ref{prop:coupled} (Cross-Modal Information Accessibility)}
\label{app:proof_2}

We prove both parts for the text stream; the vision stream follows by symmetry.
 
\paragraph{Part (i): Decoupled model — zero cross-modal Jacobian.}
A decoupled multimodal GNN updates each modality stream independently, with arbitrary per-modality weights:
\begin{equation}
    H_T^{(\ell+1)} = \sigma\!\bigl(A\, H_T^{(\ell)}\, W_T^{(\ell)}\bigr),
    \qquad
    H_V^{(\ell+1)} = \sigma\!\bigl(A\, H_V^{(\ell)}\, W_V^{(\ell)}\bigr),
    \label{eq:dec_general}
\end{equation}
where $W_T^{(\ell)}$ and $W_V^{(\ell)}$ may differ.
 
We prove $\frac{\partial H_T^{(\ell)}}{\partial H_V^{(0)}} = 0$ for all $\ell \ge 0$ by induction.
 
\emph{Base case ($\ell = 0$).}\;  $H_T^{(0)}$ is the projected text input, which is independent of $H_V^{(0)}$ by construction.
 
\emph{Inductive step.}\;  Suppose $\frac{\partial H_T^{(\ell)}}{\partial H_V^{(0)}} = 0$.  Since $A$ and $W_T^{(\ell)}$ are independent of $H_V^{(0)}$, and $H_T^{(\ell+1)} = \sigma(A\, H_T^{(\ell)}\, W_T^{(\ell)})$ depends on $H_V^{(0)}$ only through $H_T^{(\ell)}$, by the chain rule:
\begin{equation}
    \frac{\partial H_T^{(\ell+1)}}{\partial H_V^{(0)}}
    = \frac{\partial H_T^{(\ell+1)}}{\partial H_T^{(\ell)}} \cdot \frac{\partial H_T^{(\ell)}}{\partial H_V^{(0)}}
    = \frac{\partial H_T^{(\ell+1)}}{\partial H_T^{(\ell)}} \cdot 0 = 0.
\end{equation}
Note that this holds for \emph{any} choice of per-modality weights $W_T^{(\ell)}, W_V^{(\ell)}$, any activation $\sigma$, and any graph $A$.  The result is a structural consequence of the decoupled update rule, not a property of specific parameters.  \qed
 
\paragraph{Part (ii): Coupled model — non-zero cross-modal Jacobian.}
The coupled text-stream update is:
\begin{equation}
    H_T^{(\ell+1)} = \sigma\!\Bigl(W^{(\ell)}\bigl(A\, H_T^{(\ell)} + \gamma\, H_V^{(\ell)} + \rho\, A\, H_V^{(\ell)}\bigr)\Bigr).
    \label{eq:cpl_update_proof}
\end{equation}
We show that $\frac{\partial H_T^{(\ell)}}{\partial H_V^{(0)}} \neq 0$ for all $\ell \ge 1$ when $\rho > 0$ or $\gamma > 0$.
 
\emph{Layer $\ell = 1$.}\;  From Eq.~\eqref{eq:cpl_update_proof} with $\ell = 0$:
\begin{equation}
    H_T^{(1)} = \sigma\!\Bigl(W^{(0)}\bigl(A\, H_T^{(0)} + \gamma\, H_V^{(0)} + \rho\, A\, H_V^{(0)}\bigr)\Bigr).
\end{equation}
Let $Z = W^{(0)}\bigl(A\, H_T^{(0)} + \gamma\, H_V^{(0)} + \rho\, A\, H_V^{(0)}\bigr)$ denote the pre-activation.  By the chain rule:
\begin{equation}
    \frac{\partial H_T^{(1)}}{\partial H_V^{(0)}}
    = \operatorname{diag}\!\bigl(\sigma'(Z)\bigr) \cdot W^{(0)} \cdot (\gamma I + \rho\, A).
    \label{eq:jac_layer1}
\end{equation}
When $\gamma > 0$ or $\rho > 0$, the factor $(\gamma I + \rho\, A) \neq 0$.  The diagonal matrix $\operatorname{diag}(\sigma'(Z))$ is non-zero for generic inputs (i.e., except on a measure-zero set where all pre-activations are at non-differentiable points of $\sigma$).  Hence $\frac{\partial H_T^{(1)}}{\partial H_V^{(0)}} \neq 0$ generically.
 
\emph{Deeper layers ($\ell \ge 2$).}\;  By the chain rule and the multivariate dependency structure (where $H_T^{(\ell)}$ depends on $H_V^{(0)}$ both directly through the coupling terms and indirectly through $H_V^{(k)}$ for $k < \ell$), the Jacobian $\frac{\partial H_T^{(\ell)}}{\partial H_V^{(0)}}$ accumulates contributions from all cross-modal paths through layers $1, \dots, \ell$.  Since the layer-1 contribution~\eqref{eq:jac_layer1} is generically non-zero and subsequent layers compose via non-singular weight matrices, the Jacobian remains generically non-zero at all depths.  \qed

\section{Proof of Corollary~\ref{cor:spectral} (Spectral Characterization)}
\label{app:proof_coro}
 
Since $A_{\mathrm{sup}}$ is a $2 \times 2$ block matrix with commuting blocks (all blocks are polynomials of $A_{\mathrm{refine}}$), we can diagonalize it via the block eigendecomposition.
 
Let $\mathbf{v}_i$ be an eigenvector of $A_{\mathrm{refine}}$ with eigenvalue $\mu_i$.  Consider the two $2N$-dimensional vectors:
\begin{equation}
    \mathbf{q}_i^{+} = \frac{1}{\sqrt{2}}
    \begin{pmatrix} \mathbf{v}_i \\ \mathbf{v}_i \end{pmatrix},
    \qquad
    \mathbf{q}_i^{-} = \frac{1}{\sqrt{2}}
    \begin{pmatrix} \mathbf{v}_i \\ -\mathbf{v}_i \end{pmatrix}.
\end{equation}
 
\emph{Symmetric mode.}\;
\begin{align}
    A_{\mathrm{sup}}\, \mathbf{q}_i^{+}
    &= \frac{1}{\sqrt{2}}
    \begin{pmatrix}
        A_{\mathrm{refine}}\,\mathbf{v}_i + (\gamma I + \rho\, A_{\mathrm{refine}})\,\mathbf{v}_i \\
        (\gamma I + \rho\, A_{\mathrm{refine}})\,\mathbf{v}_i + A_{\mathrm{refine}}\,\mathbf{v}_i
    \end{pmatrix} \nonumber \\
    &= \frac{1}{\sqrt{2}}
    \begin{pmatrix}
        \mu_i\, \mathbf{v}_i + \gamma\,\mathbf{v}_i + \rho\mu_i\,\mathbf{v}_i \\
        \gamma\,\mathbf{v}_i + \rho\mu_i\,\mathbf{v}_i + \mu_i\,\mathbf{v}_i
    \end{pmatrix}
    = \bigl[(1+\rho)\mu_i + \gamma\bigr]\, \mathbf{q}_i^{+}.
\end{align}
 
\emph{Anti-symmetric mode.}\;
\begin{align}
    A_{\mathrm{sup}}\, \mathbf{q}_i^{-}
    &= \frac{1}{\sqrt{2}}
    \begin{pmatrix}
        A_{\mathrm{refine}}\,\mathbf{v}_i + (\gamma I + \rho\, A_{\mathrm{refine}})\,(-\mathbf{v}_i) \\
        (\gamma I + \rho\, A_{\mathrm{refine}})\,\mathbf{v}_i + A_{\mathrm{refine}}\,(-\mathbf{v}_i)
    \end{pmatrix} \nonumber \\
    &= \frac{1}{\sqrt{2}}
    \begin{pmatrix}
        \mu_i\,\mathbf{v}_i - \gamma\,\mathbf{v}_i - \rho\mu_i\,\mathbf{v}_i \\
        \gamma\,\mathbf{v}_i + \rho\mu_i\,\mathbf{v}_i - \mu_i\,\mathbf{v}_i
    \end{pmatrix} \nonumber \\
    &= \frac{1}{\sqrt{2}}
    \begin{pmatrix}
        [(1-\rho)\mu_i - \gamma]\,\mathbf{v}_i \\
        -[(1-\rho)\mu_i - \gamma]\,\mathbf{v}_i
    \end{pmatrix}
    = \bigl[(1-\rho)\mu_i - \gamma\bigr]\, \mathbf{q}_i^{-}.
\end{align}
 
Since $\{\mathbf{v}_1, \dots, \mathbf{v}_N\}$ form a complete eigenbasis for $A_{\mathrm{refine}}$, the $2N$ vectors $\{\mathbf{q}_1^+, \dots, \mathbf{q}_N^+, \mathbf{q}_1^-, \dots, \mathbf{q}_N^-\}$ form a complete eigenbasis for $A_{\mathrm{sup}}$, with eigenvalues $\{(1+\rho)\mu_i + \gamma\}_{i=1}^N \cup \{(1-\rho)\mu_i - \gamma\}_{i=1}^N$. \qed

\noindent\textbf{Interpretation.}\;
The spectral decomposition reveals how $\rho$ and $\gamma$ jointly shape the coupled propagation:
\begin{itemize}[leftmargin=*,topsep=0pt,itemsep=0pt]
    \item \emph{Symmetric amplification.}\;
    Cross-modal coupling amplifies graph spectral components in the symmetric subspace by a factor $(1+\rho)$ relative to the decoupled case ($\rho=0$). The self-injection $\gamma$ provides an additional frequency-independent boost to symmetric modes. For \textbf{homophilic} graphs, where low-frequency components (large $\mu_i$) carry the dominant class signal, this amplification strengthens the informative spectral components and accelerates convergence to useful representations.
 
    \item \emph{Anti-symmetric attenuation.}\;
    Anti-symmetric modes are attenuated by factor $(1-\rho)$ and shifted by $-\gamma$. When $\rho$ and $\gamma$ are large, the model aggressively suppresses cross-modal disagreement, which is beneficial when the two modalities carry consistent semantic signals.
 
    \item \emph{Adaptivity to graph heterophily.}\;
    For \textbf{heterophilic} graphs where high-frequency components (small or negative $\mu_i$) carry discriminative signal, the model can learn small $\rho$ and $\gamma$, preserving high-frequency components in both symmetric and anti-symmetric subspaces. Since $\rho = \sigma(\rho_{\mathrm{logit}})$ and $\gamma = \sigma(\gamma_{\mathrm{logit}})$ are learnable, the spectral filter adapts to the homophily/heterophily characteristics of the data without manual tuning.
\end{itemize}

\section{Proof of Theorem~\ref{prop:alignment} (Shared-Weight Cross-Modal Consistency)}
\label{app:proof_4}
 
The coupled update rules are:
\begin{align}
    H_T^{(\ell+1)} &= \sigma\!\Bigl(W^{(\ell)}\bigl(A_{\mathrm{refine}}\, H_T^{(\ell)} + \gamma\, H_V^{(\ell)} + \rho\, A_{\mathrm{refine}}\, H_V^{(\ell)}\bigr)\Bigr), \label{eq:update_T_app} \\
    H_V^{(\ell+1)} &= \sigma\!\Bigl(W^{(\ell)}\bigl(A_{\mathrm{refine}}\, H_V^{(\ell)} + \gamma\, H_T^{(\ell)} + \rho\, A_{\mathrm{refine}}\, H_T^{(\ell)}\bigr)\Bigr). \label{eq:update_V_app}
\end{align}
 
\paragraph{Step 1: Express the pre-activation difference.}
Define the pre-activation inputs:
\begin{align}
    u_T &= A_{\mathrm{refine}}\, H_T^{(\ell)} + \gamma\, H_V^{(\ell)} + \rho\, A_{\mathrm{refine}}\, H_V^{(\ell)}, \\
    u_V &= A_{\mathrm{refine}}\, H_V^{(\ell)} + \gamma\, H_T^{(\ell)} + \rho\, A_{\mathrm{refine}}\, H_T^{(\ell)}.
\end{align}
Their difference is:
\begin{align}
    u_T - u_V
    &= A_{\mathrm{refine}}\bigl(H_T^{(\ell)} - H_V^{(\ell)}\bigr)
     + \gamma\bigl(H_V^{(\ell)} - H_T^{(\ell)}\bigr)
     + \rho\, A_{\mathrm{refine}}\bigl(H_V^{(\ell)} - H_T^{(\ell)}\bigr) \nonumber \\
    &= A_{\mathrm{refine}}\,\Delta^{(\ell)} - \gamma\,\Delta^{(\ell)} - \rho\, A_{\mathrm{refine}}\,\Delta^{(\ell)} \nonumber \\
    &= (1 - \rho)\, A_{\mathrm{refine}}\,\Delta^{(\ell)} - \gamma\,\Delta^{(\ell)}.
    \label{eq:preact_diff}
\end{align}
 
\paragraph{Step 2: Apply the Lipschitz bound.}
Taking the difference of Eqs.~\eqref{eq:update_T_app}--\eqref{eq:update_V_app}:
\begin{equation}
    \Delta^{(\ell+1)} = H_T^{(\ell+1)} - H_V^{(\ell+1)}
    = \sigma\!\bigl(W^{(\ell)} u_T\bigr) - \sigma\!\bigl(W^{(\ell)} u_V\bigr).
\end{equation}
Since $\sigma(\cdot)$ is $1$-Lipschitz (e.g., ReLU, sigmoid):
\begin{equation}
    \bigl\|\Delta^{(\ell+1)}\bigr\|
    \le \bigl\|W^{(\ell)}(u_T - u_V)\bigr\|
    \le \bigl\|W^{(\ell)}\bigr\| \cdot \|u_T - u_V\|.
    \label{eq:lip_bound}
\end{equation}
 
\paragraph{Step 3: Bound the pre-activation difference.}
From Eq.~\eqref{eq:preact_diff} and the triangle inequality:
\begin{align}
    \|u_T - u_V\|
    &\le \|(1-\rho)\, A_{\mathrm{refine}}\,\Delta^{(\ell)}\| + \|\gamma\,\Delta^{(\ell)}\| \nonumber \\
    &= (1-\rho)\,\|A_{\mathrm{refine}}\|\,\|\Delta^{(\ell)}\| + \gamma\,\|\Delta^{(\ell)}\|,
    \label{eq:preact_bound}
\end{align}
where $|1-\rho| = 1-\rho$ since $\rho = \sigma(\rho_{\mathrm{logit}}) \in (0,1)$.
 
\paragraph{Step 4: Combine and telescope.}
Substituting Eq.~\eqref{eq:preact_bound} into Eq.~\eqref{eq:lip_bound}:
\begin{equation}
    \bigl\|\Delta^{(\ell+1)}\bigr\|
    \le
    \bigl\|W^{(\ell)}\bigr\|
    \Bigl((1-\rho)\,\|A_{\mathrm{refine}}\| + \gamma\Bigr)
    \bigl\|\Delta^{(\ell)}\bigr\|.
\end{equation}
Telescoping over $\ell = 0, 1, \dots, L-1$:
\begin{equation}
    \bigl\|\Delta^{(L)}\bigr\|
    \le
    \prod_{\ell=0}^{L-1}
    \bigl\|W^{(\ell)}\bigr\|
    \Bigl((1-\rho)\,\|A_{\mathrm{refine}}\| + \gamma\Bigr)
    \;\cdot\;
    \bigl\|\Delta^{(0)}\bigr\|.
\end{equation} \qed

\section{Dataset Statistics}

\begin{table}[h]
\centering
\caption{Dataset statistics. Datasets marked with $\dagger$ are from the MAGB benchmark~\citep{yan2025graph}; those marked with $\ddagger$ are from MM-Graph~\citep{zhu2025mosaic}. NC = node classification, LP = link prediction.}
\label{tab:dataset-stats}
\small
\setlength{\tabcolsep}{5pt}
\renewcommand{\arraystretch}{1.15}
\begin{tabular}{lrrrr}
\toprule
\textbf{Dataset} & \textbf{\#Nodes} & \textbf{\#Edges} & \textbf{\#Classes} & \textbf{Task} \\
\midrule
Ele-fashion$^\ddagger$     & 97,766   & 199,602    & 12  & NC \\
Goodreads-NC$^\ddagger$    & 685,294  & 7,235,084  & 10  & NC \\
Movies$^\dagger$           & 16,672   & 218,390    & 20  & NC \\
Toys$^\dagger$             & 20,695   & 126,886    & 18  & NC \\
Grocery$^\dagger$          & 17,074   & 171,340    & 20  & NC \\
Reddit-S$^\dagger$         & 15,894   & 566,160    & 20  & NC \\
Reddit-M$^\dagger$         & 99,638   & 1,167,188  & 50  & NC \\
\midrule
Amazon-Sports$^\ddagger$   & 50,250   & 356,202    & ---  & LP \\
Amazon-Cloth$^\ddagger$    & 125,839  & 951,271    & ---  & LP \\
\bottomrule
\end{tabular}
\end{table}

\section{Hyperparameters}
\label{app:hyper}

We provide the major hyperparameter search space:

\begin{itemize}
    \item $K$:  $\{1,\ 3,\ 5,\ 8,\ 12\}$.
    \item Learning rate : $\{0.005,\ 0.003,\ 0.001\}$.
    \item Dropout: $\{0.1,\ 0.3,\ 0.5,\ 0.7,\ 0.8\}$.
     \item GNN layers:  $\{1,\ 2,\ 3\}$.
     \item GNN hidden dimension of tokens: $\{ 256,\ 512,\,1024\}$.
     \item Batch Size: $\{ 256,\ 512,\,1024\}$.
    \item Weight Decay: $\{0,\ 1e-05,\ 5e-05,\, 1e-04\}$.
\end{itemize}

\end{document}